\documentclass{article}

\usepackage{iclr2027_conference,times}
\usepackage{amsmath,amsfonts,bm}

\def\eqref#1{equation~\ref{#1}}
\def\1{\bm{1}}

\DeclareMathAlphabet{\mathsfit}{\encodingdefault}{\sfdefault}{m}{sl}
\SetMathAlphabet{\mathsfit}{bold}{\encodingdefault}{\sfdefault}{bx}{n}

\renewcommand{\eqref}[1]{(\ref{#1})}

\usepackage{bm}
\usepackage{url}
\usepackage{hyperref}
\usepackage{booktabs}
\usepackage{tabularx}
\usepackage{graphicx}
\usepackage{amsmath,amssymb,amsthm}
\usepackage{xcolor}
\usepackage{fontawesome5}
\usepackage{tikz}
\newcommand{\paperbadge}[1]{\tikz[baseline=(badge.base)]\node[draw=black!60, fill=black!4, rounded corners=3pt, inner xsep=6pt, inner ysep=3pt, font=\small\sffamily] (badge) {#1};}

\newcommand{\nul}[1]{\textcolor{black!45}{#1}}

\newcommand{\model}[1]{\textsf{#1}}

\newcommand{\ControlJointOverFrozen}{7.9}

\newcommand{\PcqSpositionVFourMse}{2.29}
\newcommand{\PcqSpositionVFourNatural}{80.1}
\newcommand{\PcqSpositionVFourFloor}{33.3}
\newcommand{\PcqSpositionVFourRaw}{77.2}
\newcommand{\DasSpositionMse}{2.27}
\newcommand{\DasSpositionNatural}{79.0}

\newcommand{\PcqStokenVFourMse}{3.26}

\newcommand{\PcqStokenVFourFloor}{100.0}

\newcommand{\DasStokenMse}{2.93}

\newcommand{\SupervisionCCuratedGreedy}{89.94}

\newcommand{\SupervisionDCuratedHardGreedy}{91.40}
\newcommand{\SupervisionDCuratedHardGreedySd}{0.5}

\newcommand{\SupervisionEShippedGreedy}{90.84}
\newcommand{\SupervisionEShippedGreedySd}{0.9}

\newcommand{\SupervisionShuffledSoftGreedy}{89.87}

\newcommand{\SupervisionShuffledHardGreedy}{90.06}

\newcommand{\SupervisionDerangementSoftGreedy}{84.24}

\newcommand{\SupervisionCouplingGain}{0.07}
\newcommand{\SupervisionCouplingGainHard}{1.34}
\newcommand{\SupervisionPoolGain}{5.63}
\newcommand{\SupervisionMatchedUpdates}{544}
\newcommand{\SupervisionMatchedBases}{108}

\newcommand{\SupervisionEligiblePool}{383}

\newcommand{\MassMcqaConditionalGreedy}{90.84}
\newcommand{\MassMcqaConditionalGreedySd}{0.9}

\newcommand{\MassMcqaReferenceGreedy}{91.09}
\newcommand{\MassMcqaReferenceGreedySd}{1.0}

\newcommand{\MassMcqaFullGreedy}{89.02}
\newcommand{\MassMcqaFullGreedySd}{0.0}

\newcommand{\MassMcqaResolution}{0.8}

\newcommand{\MassArithmeticConditionalGreedy}{45.31}
\newcommand{\MassArithmeticConditionalGreedySd}{0.2}
\newcommand{\MassArithmeticConditionalDeclared}{73.1}

\newcommand{\MassArithmeticFullGreedy}{45.94}
\newcommand{\MassArithmeticFullGreedySd}{0.1}
\newcommand{\MassArithmeticFullDeclared}{68.9}

\newcommand{\MassArithmeticLayer}{20}

\newcommand{\ControlTermCost}{1.9}
\newcommand{\ArithControlChance}{50}
\newcommand{\ArithControlJointConditionalMass}{0.291}

\newcommand{\ArithControlJointFullMass}{0.661}
\newcommand{\ArithControlMassTrade}{15.6}

\newcommand{\ControlJointConditionalGreedy}{94.8}

\newcommand{\ControlJointConditionalError}{0.10}

\newcommand{\ControlFrozenDasGreedy}{81.7}

\newcommand{\ControlFrozenPcaGreedy}{83.9}

\newcommand{\ControlFrozenRandomOneGreedy}{34.2}

\newcommand{\ControlNoChangeError}{0.44}

\newcommand{\ControlDonorError}{0.18}

\newcommand{\ControlCap}{104.4}

\newcommand{\ArithControlJointConditionalDeclared}{93.2}
\newcommand{\ArithControlJointConditionalDeclaredSd}{4.3}
\newcommand{\ArithControlJointConditionalGreedy}{63.7}
\newcommand{\ArithControlJointConditionalGreedySd}{2.6}

\newcommand{\ArithControlJointFullDeclared}{77.3}
\newcommand{\ArithControlJointFullGreedy}{75.0}

\newcommand{\ArithControlDonorGreedy}{50.8}

\newcommand{\ArithControlCap}{208.6}
\newcommand{\ArithControlCapAmbient}{373.0}
\newcommand{\LocalisationRuns}{1188}
\newcommand{\LocalisationRunsLow}{52}
\newcommand{\LocalisationRunsHigh}{96}

\newcommand{\PsiArithmeticGemmaXCarryMean}{38}
\newcommand{\PsiArithmeticGemmaXCarryBest}{46.3}
\newcommand{\PsiArithmeticGemmaXCarryLayer}{19}

\newcommand{\PsiArithmeticGemmaXCarryDeclaredBest}{71.3}

\newcommand{\DasArithmeticGemmaXCarryMean}{31}

\newcommand{\PsiArithmeticLlamaXCarryMean}{52}

\newcommand{\DasArithmeticLlamaXCarryMean}{54}

\newcommand{\BudgetJoint}{97.5}
\newcommand{\BudgetFrozenEight}{89.6}
\newcommand{\BudgetFrozenThreeTwo}{90.4}
\newcommand{\BudgetFrozenSixFour}{90.2}

\newcommand{\IoiRandomPositionMse}{5.87}
\newcommand{\IoiRandomPositionIia}{40.5}
\newcommand{\IoiRandomTokenMse}{3.69}

\newcommand{\IoiNochangePositionMse}{16.19}
\newcommand{\IoiNochangeTokenMse}{5.39}
\newcommand{\IoiPublishedDasPositionMse}{2.20}
\newcommand{\IoiTrainedPositionMseLow}{2.20}
\newcommand{\IoiTrainedPositionMseHigh}{2.29}
\newcommand{\IoiPublishedDasTokenMse}{2.08}
\newcommand{\IoiTrainedTokenMseLow}{2.08}
\newcommand{\IoiTrainedTokenMseHigh}{3.26}

\newif\ifrev
\revfalse
\newcommand{\revcolor}{\ifrev\color{blue}\fi}
\newcommand{\rev}[1]{{\revcolor#1}}

\newcommand{\RbFilterGemmaFilteredSoft}{90.24}

\newcommand{\RbFilterGemmaUnfilteredSoft}{89.34}

\newcommand{\RbFilterGemmaSoftDrop}{0.9}
\newcommand{\RbFilterQwenSoftDrop}{1.8}

\newcommand{\RbFilterGemmaPool}{383}
\newcommand{\RbFilterGemmaPoolAll}{440}

\newcommand{\RbFilterQwenPool}{360}
\newcommand{\RbFilterQwenPoolAll}{440}

\newcommand{\RbBudgetUpdates}{768}
\newcommand{\RbBudgetExamples}{12{,}256}
\newcommand{\RbBudgetDasPairs}{251}

\newcommand{\RbBudgetPsiObservations}{383}
\newcommand{\RbOfficialPairs}{122}
\newcommand{\RbOfficialPairsPosition}{50}
\newcommand{\RbOfficialPairsSymbol}{35}
\newcommand{\RbOfficialPairsBoth}{37}

\newcommand{\RbLiteralBothPairs}{44}
\newcommand{\RbLiteralCasefoldLearnedCopy}{0.0}
\newcommand{\RbLiteralCasefoldPairs}{150}
\newcommand{\RbLiteralDigitsFullcopyCopy}{44.0}
\newcommand{\RbTransferPairs}{150}
\newcommand{\RbTransferLowercaseMin}{100.0}

\newcommand{\RbTransferDigitDonorLow}{92.4}
\newcommand{\RbTransferDigitDonorHigh}{95.8}
\newcommand{\RbTransferDigitRecipientLow}{45.3}
\newcommand{\RbTransferDigitRecipientHigh}{49.3}

\newcommand{\RbNativeDigits}{95.0}

\newcommand{\RbNativePrompts}{200}
\newcommand{\RbArcPairs}{589}
\newcommand{\RbArcWrongDonors}{111}
\newcommand{\RbArcTestQuestions}{1178}
\newcommand{\RbArcExcluded}{4}

\newcommand{\RbArcSoftOfficial}{70.03}

\newcommand{\RbArcSoftAgreement}{94.8}
\newcommand{\RbArcSoftAgreementWrong}{85.9}

\newcommand{\RbArcSoftKl}{0.020}

\newcommand{\RbArcConfidenceAgreementWrong}{84.7}

\newcommand{\RbArcHardAgreementWrong}{79.3}

\newcommand{\RbArcHardKl}{0.117}

\newcommand{\RbArcGoldOfficial}{71.45}

\newcommand{\RbArcGoldAgreementWrong}{72.1}

\newcommand{\RbArcGoldKl}{0.090}

\newcommand{\RbArcValidationPairs}{283}
\newcommand{\RbArcValidationQuestions}{566}
\newcommand{\RbArcTrainQuestions}{2241}
\newcommand{\RbArcTrainAccuracy}{79.7}
\newcommand{\RbArcTopCorrect}{0.78}

\newcommand{\RbArcTopConfident}{28.7}
\newcommand{\RbArcUpdates}{768}

\title{Programmable Semantic Interventions:\\ Learning Causal Carriers from Behavioral Programs}

\author{%
    Paolo Muratore\\
    EPFL\\
    \texttt{paolo.muratore@epfl.ch} \\
    \And
    Mackenzie Weygandt Mathis\\
    EPFL\\
    \texttt{mackenzie.mathis@epfl.ch}
}

\iclrfinalcopy

\begin{document}

\maketitle
\lhead{} %

\begin{abstract}
Causal abstraction explains a frozen network by aligning its activations with the variables of a simpler, interpretable model, and checking that interventions on the two agree. Finding such an alignment currently requires a counterfactual dataset: for every pair of prompts, the answer the network ought to give once the variable has been changed. \rev{In practice, those labels are produced by }an executable algorithm for the behavior\rev{, written }before studying the model supposed to implement it. However, the criterion these methods optimize only leverages the answer an intervention should produce. Here, we develop an approach that lets the investigator define that relationship directly: \rev{the answers the variable can produce, and which of them an intervention should yield}. We call this interface a \emph{semantic program} and introduce \emph{Programmable Semantic Interventions} (PSI, $\Psi$): a low-rank activation subspace learned so that interchanging it \rev{changes the model's answer as the program declares, with the expected answers read from the frozen model itself}. On the MIB benchmark $\Psi$ \rev{performs on par with supervised alignment at matched data and training budget, without curated pairs, and it needs no ground-truth labels; while on natural science questions with gold labels withheld, it tracks the model's own answers better than a subspace trained on gold labels}. The declaration also fixes a coordinate system with known semantics, enabling donor-free steering: jointly training $\Psi$ for both localization and control realizes $\ControlJointConditionalGreedy$\% of held-out requests for answers the investigator chooses, without the need for donor prompts.
\end{abstract}

\ificlrfinal
\begin{center}
\hypersetup{pdfborder={0 0 0}}%
\href{https://github.com/AdaptiveMotorControlLab/psi}{\paperbadge{\raisebox{-0.12ex}{\faCode}\, Code}}%
\hspace{0.7em}%
\href{https://adaptivemotorcontrollab.github.io/psi/}{\paperbadge{\raisebox{-0.12ex}{\faGlobeAmericas}\, Website}}%
\end{center}
\fi

\section{Introduction}
\label{sec:introduction}

Mechanistic interpretability sets out to explain the inner workings of a network in terms of the algorithm it implements \citep{olah2020zoom, sharkey2025open}. Recently, an increased emphasis has been placed on providing \emph{causal} claims \citep{mueller2024quest}, i.e., what part of a
network's activations encodes an abstract, high-level (simpler) variable, and does an intervention move the network's behavior in a predictable way that the corresponding high-level model says it should \citep{geiger2025causal}? The latter distinction is essential: reading a property off an activation with a trained probe establishes that the property is present \citep{belinkov2022probing}, while a causal claim asks whether the network actually uses it. Namely, causal abstraction makes the causal claim testable \citep{geiger2021causal, geiger2025causal}, and standard benchmarks now exist to evaluate methods against it \citep{arora2024causalgym, mueller2025mib}. However, testing \textit{causal claims} at scale currently requires the construction of a counterfactual dataset. For every pair of inputs differing (only) in the value of a putative high-level variable, one must build the answer the network ought to give once that value has been exchanged. Methods in the distributed alignment lineage obtain that answer by executing a high-level algorithm the investigator has specified in advance \citep{geiger2024finding, wu2023boundless, huang2024ravel, sun2025hyperdas}. \rev{In current practice, the investigator therefore }commits to an algorithm for the behavior before studying the model that is supposed to implement it, and the approach reaches only those behaviors for which such an algorithm can be written down, using \rev{a curated list of prompt pairs}.

The success criterion for identification is \emph{interchange intervention accuracy} \citep{geiger2022inducing}, and current methods are directly optimized against it. However, the criterion itself never inspects the high-level algorithm: it merely relies on the answer the algorithm produces under the intervention and compares it with the intervened network's output. This simple observation points to a significant gap between the amount of required specifications and what is ultimately used. Concretely, consider the problem of localizing the variable encoding the position of the correct option in a four-choice question. A full high-level causal specification, \rev{as used in current practice}, is an executable algorithm that \rev{finds the option matching the question, records its position and outputs the label found there}. We propose a different interface -- semantic programs -- \rev{in which the investigator writes only the possible answers and the rule that gives the expected one}.

A \emph{semantic program} \rev{lists }the answers a variable \rev{can produce and states which of them an intervention should yield; the expected answer itself }is measured from the frozen model. We call the method built on it \emph{Programmable Semantic Interventions} (PSI, $\Psi$): a low-rank activation subspace, learned so that interchanging it \rev{between two prompts makes the first answer as the program prescribes}. Moreover, a program written in the model's own answer space induces a coordinate system whose semantics the investigator knows, such that an arbitrary behavior can be directly \emph{requested} in it. Coming back to the question-answering example, a semantic program for the same task consists of just two lines: a four-option library of labels, plus the rule that an interchange must replace the base's label with the source's. \rev{As such, it requires no ground-truth labels, and it can be used to request a reading that no prompt in the data exhibits.}

\textbf{Related work.} Work on localizing causal variables has steadily reduced the choices that an investigator makes by hand. Distributed alignment search learns the subspace by gradient descent \citep{geiger2024finding}, its boundless variant learns the rank along with it \citep{wu2023boundless}, multi-task training adds attribute isolation \citep{huang2024ravel}, and a hypernetwork takes over the search across token positions \citep{sun2025hyperdas}. However, throughout this progression, the requirement of an executable high-level model remains, and each method trains on a curated list of prompt pairs. $\Psi$ thus follows this lineage by \rev{changing where the training labels come from }rather than adding more automation machinery. A second line of work encodes a behavior into activations with no donor example. Directions fitted from contrastive prompts steer a model at inference \citep{zou2023representation, turner2023steering, rimsky2024steering, li2023inference}, and recent methods learn the write operator \citep{bao2026faithful, rodriguez2025lineas} or amortize it over a space of requests \citep{shi2026unisteer}.  
Localizing a variable and controlling it are known to be distinct capabilities: the directions that read a concept and the directions that change it are geometrically distinct \citep{park2024linear, galeone2026perfect, wu2025axbench}, steering vectors are brittle across prompts \citep{tan2024analysing}, and a subspace can move an output through a pathway unrelated to the feature it names \citep{makelov2024subspace}. We thus validate both capabilities in Section~\ref{sec:results} and refer to Appendix~\ref{sec:appendix-related} for a broader overview of the relevant literature.

\textbf{Contributions.} We introduce the semantic program as the interface that a causal-variable method requests from the investigator, and $\Psi$, which learns a carrier from it. \rev{Four }results follow: (\emph{i}) on the causal-variable localization tasks, $\Psi$ reaches the baseline's quality without a counterfactual dataset, and \rev{performs on par with it at matched data and training budget}; (\emph{ii}) ablating the supervision, we measure the value of a counterfactual dataset and find that a method requiring no pair list recovers most of the performance by benefiting from a larger donor pool size which provides observations the pair list cannot reach; (\emph{iii}) \rev{because the labels are read from the model, $\Psi$ needs no ground truth: on natural science questions with gold labels withheld, it tracks the model's own answers, wrong ones included, better than a subspace trained on gold labels; (}\emph{\rev{iv}}\rev{) finally, because the answer list is fixed in advance}, the same carrier realizes readings that the investigator requests, with no source prompt and including readings that no prompt in the data exhibits, surpassing what an interchange from a real donor achieves at the same site.

\section{Methods}
\label{sec:methods}
\begin{figure}[t]
    \centering
\includegraphics[width=0.95\textwidth]{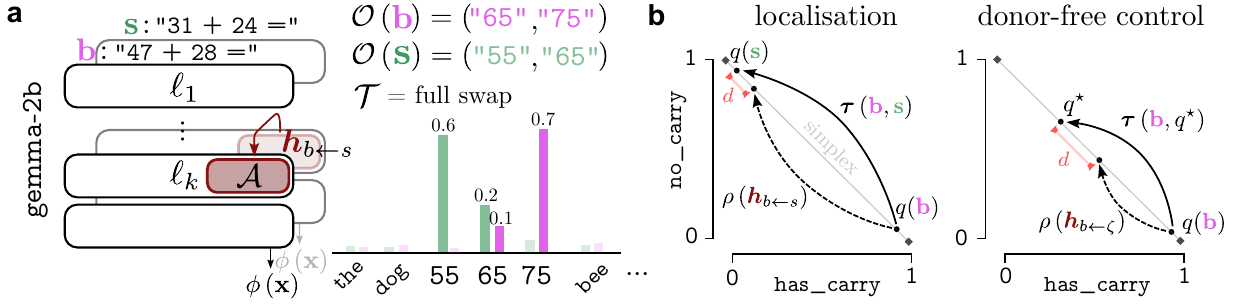}
    \caption{\textbf{Programmable Semantic Interventions.} \textbf{(a)} The
    investigator declares which answers $\mathcal{O}$ the variable is about: here the two numerical sums a carry does and does not produce, which result in different strings for the base $\mathbf{b}$ and the source $\mathbf{s}$. The frozen subject supplies its probability on each of them, read at the cut $\ell_k$ where the abstraction $\mathcal{A}$ selects the carrier. \textbf{(b)} A reading is a point in the simplex over that library. Interchanging the source's carrier should move the base's reading to $\bm{\tau}(\mathbf{b},\mathbf{s})$, and the distance $d$ to what the intervened model reads is the training signal; the same target can be named directly as $q^\star$ and realized by an edit, with no source prompt.}
    \label{fig:teaser}
\end{figure}

\textbf{Problem setup.} We consider a frozen subject $\phi$, an input prompt $\mathbf{x}$, and indicate with
$\phi^n(\mathbf{x})$ the standard auto-regressive greedy rollout of the
model predictions for a sentence length $n$. We indicate by
$\bm{h} = \phi|_\kappa(\mathbf{x})$ the hidden vector of activations $\bm{h} \in \mathbb{R}^{d}$ produced by the model, where $\kappa$ is a set of location indices (e.g.\ the collection of token positions and layers in the residual stream of a transformer network). \emph{Causal abstraction} \citep{geiger2025causal} formalizes what it means for such a vector to carry a concept: an abstraction $\mathcal{A}$ maps the low-level states of the network to the states of a high-level model $\mathbf{H}$, whose variables are the concepts we wish to reason with, and it is faithful under a commutative condition on interventions—intervening on the low-level model and abstracting the result should agree with abstracting first and then performing the \emph{corresponding} intervention on $\mathbf{H}$ (see \citealp{geiger2025causal} for the formal exposition). In our experiments, $\mathcal{A} : \mathbb{R}^{d} \to \mathbb{R}^{r}$ is a linear map from the hidden vector $\bm{h}$ to an $r$-dimensional high-level representation $\bm{z} = \mathcal{A}(\bm{h})$, and an \emph{intervention} is performed by replacing part of the hidden vector $\bm{h}_b = \phi|_\kappa(\mathbf{b})$ of a base prompt $\mathbf{b}$ with the corresponding part of the hidden vector $\bm{h}_s = \phi|_\kappa(\mathbf{s})$ of a source prompt $\mathbf{s}$, and then measuring the resulting output $\phi^n_{b \leftarrow s}(\mathbf{b})$.

\textbf{Measuring abstraction quality.} Empirically, one quantifies the faithfulness of a candidate abstraction $\mathcal{A}$ by measuring the agreement between the two sides of the commutative condition over a distribution of interventions. An intervention is a pair $\iota = (\mathbf{b}, \mathbf{s})$ of a base and a source prompt, drawn from $\mathcal{B} \times \mathcal{S}$. The \emph{interchange intervention accuracy} (IIA) \citep{geiger2022inducing, geiger2024finding} has been proposed as one such criterion: it measures how often the intervened low-level model and the high-level model produce the same output\rev{:
}

\begin{equation}
  \mathrm{IIA}
  \;=\;
  \mathbb{E}_{\iota \sim \mathcal{B} \times \mathcal{S}}
  \Big[\,
    \1\big[\,
      \phi^n_{b \leftarrow s}(\mathbf{b})
      \;=\;
      \mathbf{H}_{b \leftarrow s}(\mathbf{b})
    \,\big]
  \,\Big],
  \label{eq:iia}
\end{equation}

where, with a slight abuse of notation, $\mathbf{H}_{b \leftarrow s}(\mathbf{b})$ denotes the high-level model's output under the intervention corresponding to $b \leftarrow s$, and the indicator $\1[\cdot]$ evaluates to $1$ when the two outputs agree and $0$ otherwise.

\textbf{Aligning a subspace.} Replacing the hidden vector wholesale would test whether the \emph{entire} activation carries the concept (the \emph{Full Vector} baseline of the MIB suite, \citealp{mueller2025mib}), which is a rather crude baseline: a single residual stream is expected to carry many concurrent variables, more of them than it has dimensions \citep{elhage2022toy}. A common alternative is to let the abstraction single out a low-dimensional subspace \citep{geiger2024finding} or a sparse set of learned features \citep{cunningham2024sparse, gao2025scaling}, leaving the rest of the activation untouched. Concretely, in our work we take an orthonormal $A \in \mathbb{R}^{r \times d}$ with $r \ll d$, write $\bm{\pi} = A^{\top}A$ for the projector onto its row space, and define the intervention that transplants the source's \emph{carrier} into the base while preserving the base's complement:

\begin{equation}
  {\bm{h}_{b \leftarrow s}} = \bm{\pi} \bm{h}_s + (\mathrm{I} - \bm{\pi}) \bm{h}_b .
  \label{eq:interchange}
\end{equation}

The abstraction is then simply the readout of that subspace, $\mathcal{A}(\bm{h}) = A \bm{h}$, and identifying the causal variable amounts to \emph{finding} $A$. \rev{In }our experiments we restrict ourselves to linear maps as non-linear alignments weaken what a successful abstraction certifies \citep{sutter2025non}. Moreover, a linear carrier keeps the intervention of Eq.~\eqref{eq:interchange} a projection, so the edit has a scale we can compare against the displacements interchange interventions naturally produce.

\textbf{What the IIA criterion requires.} One observation about Eq.~\eqref{eq:iia} motivates our interface proposal (Section \ref{sec:psi}). The IIA criterion never inspects the inner workings of $\mathbf{H}$. It reads a single quantity off the high-level model: the output it produces under the intervention, while remaining oblivious to its variables, its mechanisms, or its causal graph. Existing methods \citep{geiger2024finding, wu2023boundless, huang2024ravel, sun2025hyperdas} nonetheless specify $\mathbf{H}$ in full, as an executable algorithm, and obtain that output by running it. The result of that computation is then taken as the counterfactual label attached to each training pair. Thus, the full specification of $\mathbf{H}$ is typically taken as a prerequisite for causal investigation. In particular, it requires postulating a particular algorithm the low-level model $\phi$ is supposed to implement before one has studied the model itself, and it confines the method to tasks for which a full algorithmic specification is available.

\subsection{Programmable Semantic Interventions}
\label{sec:psi}

Rather than specifying the high-level model $\mathbf{H}$ fully and computing its output, the investigator \emph{declares} that output directly, as a rule stated in the subject's own answer space. This declaration requires two components: an outcome library $\mathcal{O}$ that specifies which answers the concept pertains to, and a transport rule $\mathcal{T}$ that indicates how those answers should be transferred under an intervention. We call the resulting pair of objects that make up this declaration a \emph{semantic program}, $\mathcal{P} = \left( \mathcal{O}, \mathcal{T} \right)$. We offer a graphical overview of the method in Figure~\ref{fig:teaser}.

\paragraph{The outcome library.} The outcome library is a small, prompt-dependent set of candidate answers, one for each value the variable of interest can take:
\begin{equation}
  \mathcal{O}(\mathbf{x}) \;=\; \big(o_1(\mathbf{x}), \dots, o_k(\mathbf{x})\big) .
  \label{eq:library}
\end{equation}

\rev{To make this concrete, consider a multiple-choice question, with the position of the correct option as the variable of interest (the MCQA task of \citealp{mueller2025mib}). The prompt $\mathbf{x}$ asks \emph{Coconuts are brown. What color are coconuts?} and lists four colors labelled }\texttt{\rev{X}}\rev{, }\texttt{\rev{C}}\rev{, }\texttt{\rev{P}} \rev{and }\texttt{\rev{A}}\rev{, with brown in second position; the library is the prompt's own labels in the order it lists them, $\mathcal{O}(\mathbf{x}) = (\texttt{"X"}, \texttt{"C"}, \texttt{"P"}, \texttt{"A"})$, indexed by the four values of the position. Writing it requires neither the correct answer nor any account of how the model finds it. For an entity-attribute lookup (e.g.\ the RAVEL task of \citealp{huang2024ravel}) the library is the set of admissible attribute values, while for the carry in two-digit addition it is the two sums with and without a carry (Figure~\ref{fig:teaser}), and writing it requires solving the task.} We highlight two properties of Eq.~\eqref{eq:library}. First, its entries are answers in the subject's vocabulary: they are the (potentially multi-token) strings the model would have to emit. Second, the library says nothing about which of the library entries is \emph{correct}. Its role is to fix a (semantic) reference frame, i.e.\ the set of outputs this concept is about and expected to elicit, and to leave the readings within that frame to the model.

Once a frame is declared, the subject $\phi$ supplies its own coordinates within it. Writing $\rho_i(\mathbf{x}) = \log p_\phi\big(o_i(\mathbf{x}) \mid \mathbf{x}\big)$ for the log-probability the model assigns to the $i$-th library entry, we define the following two quantities:
\begin{equation}
  Z(\mathbf{x}) = \sum_{i=1}^{k} e^{\rho_i(\mathbf{x})},
  \qquad
  q_i(\mathbf{x}) = \frac{e^{\rho_i(\mathbf{x})}}{Z(\mathbf{x})} .
  \label{eq:measurement}
\end{equation}
The \emph{declared mass} $Z(\mathbf{x}) \in [0,1]$ \rev{(the entries are distinct and none is a prefix of another, so they name disjoint continuations) }answers how much of the model's probability the frame captures at all. $Z (\mathbf{x})$ thus quantifies the coverage of the declared library. The \emph{reading} $q(\mathbf{x}) = [q_1 (\mathbf{x}), \dots, q_k (\mathbf{x})]$ is how the captured mass divides between the declared alternatives, and it is the vector we treat as the model's opinion about the variable. Continuing the \rev{multiple-choice} example, a subject that places $p = 0.70$ on \texttt{\rev{"C"}} and $p = 0.05$ on \rev{each of the other three labels }has $Z(\mathbf{x}) = \rev{0.85}$ and $q (\mathbf{x}) = \rev{(0.06, 0.82, 0.06, 0.06)}$: \rev{most }of its mass lies inside the frame, and within the frame it is confident the \rev{correct option is the second}. Both numbers are thus measured from the frozen model and require no further investigator input.

\paragraph{The transport rule.} The second half of the program states how a reading $q (\mathbf{x})$ should change when the variable is intervened upon. We represent this transformation as a map:
\begin{equation}
  \bm{\tau} = \mathcal{T} \left( q (\mathbf{s}), q(\mathbf{b}), \gamma \right) ,
  \label{eq:transport_rule}
\end{equation}
which takes the source's reading $q (\mathbf{s})$, the base's reading $q(\mathbf{b})$, and an optional per-example gate $\gamma$, and returns the reading the base \emph{should} exhibit once the source's value of the variable has been written into it. The source reading $q (\mathbf{s})$ can be measured by a declared query for the variable or a donor prompt (see Appendix~\ref{sec:appendix-programs}).

The simplest and most common rule is the full swap, $\bm{\tau} = q(\mathbf{s})$. Return to the \rev{multiple-choice} example, and let the base \rev{$\mathbf{b}$ be the coconut question (second position) and the source $\mathbf{s}$ a question whose correct option is the fourth, with labels }\texttt{\rev{K}}\rev{, }\texttt{\rev{B}}\rev{, }\texttt{\rev{Q}} \rev{and }\texttt{\rev{D}}. If $A$ really is the subspace that carries the \rev{position}, then transplanting the source's carrier into the base should make the base behave as though \rev{the correct option were the fourth}: it should answer \texttt{\rev{A}} rather than \texttt{\rev{C}}. Note what makes this well posed: the two prompts have different libraries, $\mathcal{O} (\mathbf{b}) = \rev{(\texttt{"X"},\texttt{"C"},\texttt{"P"},\texttt{"A"})}$ against $\mathcal{O} (\mathbf{s}) = \rev{(\texttt{"K"},\texttt{"B"},\texttt{"Q"},\texttt{"D"})}$, but both are indexed by the same variable, so a reading transports even though a token does not.

However, not every variable is decisive in every context. Consider an entity-attribute task, where the prompt selects which attribute is being asked about the introduced entity. In such a task, interventions should alter the model's behavior only when the base prompt queries the sought attribute (e.g.\ in the RAVEL task, intervening on the entity's continent should change the answer to a continent question while leaving a language question untouched). Such cases can be easily covered by a gated rule, $\bm{\tau} = \gamma (\mathbf{b}) q(\mathbf{s}) + (1 - \gamma (\mathbf{b})) q(\mathbf{b})$, where $\gamma (\mathbf{b}) \in \{0,1\}$ records whether the intervened variable is the operative one for that prompt; the target is the source's reading when it is, and the base's own reading, i.e.\ a no-op, when it is not. More generally, $\mathcal{T}$ may be given as a table specifying, for each pair of source and base values, the reading that should result. We refer to Appendix~\ref{sec:appendix-programs} for a complete description of the programs used in our experiments.

\paragraph{Semantic programs are \rev{label-free}.} A semantic program $\mathcal{P}$ is written before any training and consists of Eqs.~\eqref{eq:library} and~\eqref{eq:transport_rule}: a family of candidate answers and a rule for moving between readings of them. It contains no correct answer, no counterfactually annotated example pair, and no executable high-level model. Everything numerical in the target of Eq.~\eqref{eq:transport_rule} is measured from the frozen subject through Eq.~\eqref{eq:measurement}. \rev{Training on a program therefore needs no ground-truth label at any stage: base and source are paired at random, and every target is read from the subject, including on prompts it answers wrongly. This lets $\Psi$ localize a variable on tasks for which no gold answer is available (Section~\ref{sec:results-natural}).} Appendix Table~\ref{tab:supervision} accounts for what each approach asks of the investigator, and Section~\ref{sec:results} measures what each input is worth.
\subsection{Learning the alignment from the program}
\label{sec:objective}

A semantic program $\mathcal{P}$ determines a training signal without any further input. Given a collection of base prompts $\mathcal{B}$, we can construct candidate base-source interventions $\iota = (\mathbf{b}, \mathbf{s})$ by pairing via a random derangement within a group of prompts the program declares comparable. The transport rule of Eq.~\eqref{eq:transport_rule} then produces a target reading $\bm{\tau} (\mathbf{b}, \mathbf{s})$, while the interchange of Eq.~\eqref{eq:interchange} produces an activation $\bm{h}_{b \leftarrow s}$ whose \rev{library log-probabilities }$\rho(\bm{h}_{b \leftarrow s})$ we can measure\rev{, by completing the subject's forward pass from the edited activation (we write $\rho(\bm{h})$, $q(\bm{h})$ and $Z(\bm{h})$ for the quantities of Eq.~}\eqref{eq:measurement} \rev{measured this way)}. Learning the subspace $A$ is then a matter of making the second agree with the first:

\begin{equation}
  \mathcal{L}_{\mathrm{donor}}(A)
  \;=\;
  \mathbb{E}_{\mathbf{b},\mathbf{s}}\;
  d \left( \rho(\bm{h}_{b \leftarrow s}),\; \bm{\tau}(\mathbf{b}, \mathbf{s})\right) .
  \label{eq:donor}
\end{equation}

Equation~\eqref{eq:donor} is a differentiable relaxation of Eq.~\eqref{eq:iia}: the indicator that asks whether two outputs match exactly is replaced by a divergence $d$ that measures how far apart two readings are, and the expectation is taken over sampled prompt pairs. One natural choice for $d$ is the cross-entropy of the target against the model's log-probabilities \citep{geiger2024finding}. Because the $\rho_i$ are log-probabilities over the whole vocabulary rather than over the library, that quantity decomposes as

\begin{equation}
  -\sum_{i} \tau_i \,\rho_i
  \;=\;
  \underbrace{H(\bm{\tau}) \;+\; \mathrm{KL}\big(\bm{\tau} \,\big\|\, q(\bm{h}_{b \leftarrow s})\big)}_{\text{the reading}}
  \;\;\underbrace{-\;\log Z(\bm{h}_{b \leftarrow s})}_{\text{the mass}} ,
  \label{eq:decomposition}
\end{equation}

where $H$ is the entropy of the target and $\mathrm{KL}$ the Kullback--Leibler divergence. The first group asks the intervened model to produce the requested reading; the second asks it to place as much probability as possible \emph{inside the library}. The \textsc{conditional} form drops the mass term and scores the reading alone; the \textsc{reference} form targets the mass $Z(\mathbf{b})$ the unintervened prompt already shows, so an intervention is asked to preserve it; the \textsc{full} form keeps Eq.~\eqref{eq:decomposition} as written. We used the \textsc{full} form for the Arithmetic localization experiments and \textsc{conditional} for the rest (Table~\ref{tab:localisation}), while Appendix~\ref{sec:appendix-mass} measures the impact each choice has on control costs.

\subsection{Enabling control via the same interface}
\label{sec:control}

A trained subspace that minimizes Eq.~\eqref{eq:donor} supports interchange interventions, which realize the reading of whichever source prompt is supplied. But one crucial advantage of the reference frame of Eq.~\eqref{eq:library} is that it represents a coordinate system with \emph{known} semantics. Such a coordinate system can thus be used to naturally formulate any representable reading target $q^\star$, while the transport rule $\mathcal{T}$ can then be used to \emph{request} such a reading from the model when exposed to a base prompt $\mathbf{b}$, i.e. construct $\bm{\tau}^\star = \mathcal{T} \left(q^\star, q(\mathbf{b}), \gamma \right)$. In particular, we can attach a controller:

\begin{equation}
  \bm{\zeta} \;=\; \mathrm{control}\big( \bm{\tau}^{\star},\; q(\mathbf{b}),\; A\bm{h}_b \big) ,
  \label{eq:controller}
\end{equation}

which maps a requested reading $\bm{\tau}^{\star}$, the subject's current reading $q(\mathbf{b})$, and the base's carrier coordinates $A\bm{h}_b$ to an edit $\bm{\zeta} \in \mathbb{R}^{r}$ expressed in those same carrier coordinates. The steered activation is then $\bm{h}_{b \leftarrow \zeta} = \bm{h}_b + A^{\top}\bm{\zeta}$: by construction the edit lives entirely inside the identified subspace and leaves the complement $(\mathrm{I} - \bm{\pi})\bm{h}_b$ untouched, exactly as the interchange of Eq.~\eqref{eq:interchange} does.

The simplest instance, and the one we use in our experiments, is a linear gain driven by the residual between the requested and the current reading. Writing $\mathrm{lo}(\bm{p}) \in \mathbb{R}^{k-1}$ for the log-odds of a reading against its last entry, $\mathrm{lo}_i(\bm{p}) = \log p_i - \log p_k$, the gain controller is:

\begin{equation}
  \bm{\zeta} \;=\; G \,\big[\, \mathrm{lo}(\bm{\tau}^{\star}) \;-\; \mathrm{lo}\big(q(\mathbf{b})\big) \,\big],
  \qquad G \in \mathbb{R}^{r \times (k-1)} ,
  \label{eq:linear_controller}
\end{equation}

so that $G$ is the only learned object and the edit vanishes whenever the request coincides with what the model is already doing. We clip the edit to a fixed budget, $\bm{\zeta} \leftarrow \bm{\zeta}\,\min\!\left(1,\, c / \lVert \bm{\zeta} \rVert\right)$, with $c$ set to a high quantile of the displacements that interchange interventions themselves produce at that site (Appendix~\ref{sec:appendix-hparams}).

The controller is trained jointly with the subspace, against requests drawn half from the readings other prompts exhibit and half uniformly from the probability simplex, so that it is exercised on targets no prompt in the dataset elicits. With $d$ the same divergence as before, the complete objective is

\begin{equation}
  \mathcal{L}(A, G)
  \;=\;
  \underbrace{\mathbb{E}_{\mathbf{b},\mathbf{s}}\;
    d\big( \rho(\bm{h}_{b \leftarrow s}),\; \bm{\tau}(\mathbf{b}, \mathbf{s}) \big)}_{\text{donor: the subspace carries the variable}}
  \;+\;
  \lambda\,
  \underbrace{\mathbb{E}_{\mathbf{b},\, \bm{\tau}^{\star}}\;
    d\big( \rho(\bm{h}_{b \leftarrow \zeta}),\; \bm{\tau}^{\star} (\mathbf{b}, q^\star) \big)}_{\text{control: the subspace is writable}} .
  \label{eq:full_loss}
\end{equation}

The two terms share the abstraction $A$, which is what makes the training joint: the donor term asks the subspace to behave like the variable under transplantation, and the control term asks the same subspace to be reachable by a learned edit. At inference, the controller of Eq.~\eqref{eq:controller} accepts any distribution over the declared library, including readings $q^\star$ no prompt in the dataset elicited, such as an even split between two answers, with no source prompt and no retraining. This is a capability methods trained on the interchange of Eq.~\eqref{eq:interchange} \citep{geiger2024finding, wu2023boundless, huang2024ravel, sun2025hyperdas} do not offer directly, since their interchange operator requires a donor\rev{. The controller is thus a steering method, which writes only inside a subspace that has first passed the interchange test}. In Section~\ref{sec:results-control} we therefore equip their frozen carriers with a controller of our own and measure what that recovers.

\section{Results}
\label{sec:results}

\textbf{Protocol.} We evaluate on the causal-variable tasks of \citet{mueller2025mib}, with the benchmark's subjects \model{gemma-2\_2b}, \model{llama-3.1\_8b} and \model{qwen-2.5\_0.5b}, and with \model{gpt2-small} for IOI. At each site we set the carrier rank to the subspace dimension the official DAS baseline uses there, so the two are compared at equal capacity, and we follow the benchmark's site-selection protocol: at each layer we take the best of the benchmark's fixed token positions, then report the mean of that depth profile together with its best layer. We also include a \emph{random} subspace of the same rank at the same sites as a floor. Localization runs use $\lambda = 0$ in Eq.~\eqref{eq:full_loss}, while the control experiments of Section~\ref{sec:results-control} use $\lambda = 1$ and focus on \model{gemma-2\_2b} at two sites. Reported localization numbers are single-seed, following \citet{mueller2025mib} (Appendix~\ref{sec:appendix-compute}); at the one site (MCQA $X_{\mathrm{Order}}$, \model{gemma-2\_2b} layer $17$, \textsf{last\_token}) where we repeated the experiment three times, the spread was $\pm \MassMcqaConditionalGreedySd$ points, comparable to one evaluation pair ($\MassMcqaResolution$ pt).

\subsection{Localizing causal variables}
\label{sec:results-localisation}

\begin{table}[t]
\centering
\footnotesize
\setlength{\tabcolsep}{4pt}
\caption{\textbf{Localizing causal variables on MIB.} Interchange intervention accuracy (\%, higher is better): the mean over the depth profile, with the best layer in parentheses. $\Psi$ uses the rank the DAS baseline uses at each cell. \textbf{Bold} is the best method in a column, mean and best layer separately, and the \nul{gray} rows are nulls that do not compete for it. Baselines from \citet{mueller2025mib}.}
\label{tab:localisation}
\begin{tabular}[t]{@{}l cc cc@{}}
\toprule
 & \multicolumn{4}{c}{ARC (Easy)} \\
\cmidrule(l){2-5}
 & \multicolumn{2}{c}{\model{gemma-2\_2b}} & \multicolumn{2}{c}{\model{llama-3.1\_8b}} \\
\cmidrule(lr){2-3}\cmidrule(l){4-5}
Method & $O_{\mathrm{Answer}}$ & $X_{\mathrm{Order}}$ & $O_{\mathrm{Answer}}$ & $X_{\mathrm{Order}}$ \\
\midrule
\nul{Full Vector} & \nul{63 (100)} & \nul{43 (74)} & \nul{68 (100)} & \nul{47 (72)} \\
\nul{Random} & \nul{18 (27)} & \nul{33 (33)} & \nul{17 (28)} & \nul{33 (33)} \\
DBM & 82 (\textbf{99}) & 63 (80) & 85 (\textbf{100}) & 69 (82) \\
+ PCA & 78 (98) & 64 (81) & 84 (\textbf{100}) & 72 (83) \\
+ SAE & 70 (89) & 54 (70) & 74 (94) & 55 (67) \\
DAS & 88 (94) & \textbf{76} (88) & 88 (99) & 74 (84) \\
\midrule
$\Psi$ (ours) & \textbf{95} (98) & 71 (\textbf{89}) & \textbf{97} (\textbf{100}) & \textbf{80} (\textbf{92}) \\
\bottomrule
\end{tabular}
\hfill
\begin{tabular}[t]{@{}l c c@{}}
\toprule
 & \multicolumn{2}{c}{Arithmetic (+)$^{\dagger}$} \\
\cmidrule(l){2-3}
 & \multicolumn{1}{c}{\model{gemma-2\_2b}} & \multicolumn{1}{c}{\model{llama-3.1\_8b}} \\
\cmidrule(lr){2-2}\cmidrule(l){3-3}
Method & $X_{\mathrm{Carry}}$ & $X_{\mathrm{Carry}}$ \\
\midrule
\nul{Full Vector} & \nul{29 (35)} & \nul{35 (45)} \\
\nul{Random} & \nul{26 (26)} & \nul{26 (26)} \\
DBM & 33 (43) & 47 (58) \\
+ PCA & 32 (44) & 37 (56) \\
+ SAE & 32 (44) & 38 (55) \\
DAS & 31 (35) & \textbf{54} (\textbf{65}) \\
\midrule
$\Psi$ (ours) & \textbf{38} (\textbf{46}) & 52 (64) \\
\bottomrule
\end{tabular}
\par\bigskip
\begin{tabular}{@{}l cc cc cc@{}}
\toprule
 & \multicolumn{6}{c}{MCQA} \\
\cmidrule(l){2-7}
 & \multicolumn{2}{c}{\model{gemma-2\_2b}} & \multicolumn{2}{c}{\model{llama-3.1\_8b}} & \multicolumn{2}{c}{\model{qwen-2.5\_0.5b}} \\
\cmidrule(lr){2-3}\cmidrule(lr){4-5}\cmidrule(l){6-7}
Method & $O_{\mathrm{Answer}}$ & $X_{\mathrm{Order}}$ & $O_{\mathrm{Answer}}$ & $X_{\mathrm{Order}}$ & $O_{\mathrm{Answer}}$ & $X_{\mathrm{Order}}$ \\
\midrule
\nul{Full Vector} & \nul{61 (100)} & \nul{44 (77)} & \nul{77 (100)} & \nul{46 (68)} & \nul{35 (99)} & \nul{49 (99)} \\
\nul{Random} & \nul{19 (28)} & \nul{33 (33)} & \nul{16 (31)} & \nul{33 (33)} & \nul{7 (18)} & \nul{33 (33)} \\
DBM & 84 (\textbf{99}) & 63 (84) & 86 (\textbf{100}) & 66 (73) & 46 (94) & 60 (99) \\
+ PCA & 57 (96) & 52 (81) & 65 (99) & 53 (74) & 22 (76) & 54 (\textbf{100}) \\
+ SAE & 73 (90) & 51 (65) & 80 (99) & 58 (65) & --- & --- \\
DAS & \textbf{95} (97) & \textbf{77} (\textbf{93}) & 94 (\textbf{100}) & 77 (\textbf{91}) & 86 (95) & \textbf{78} (\textbf{100}) \\
\midrule
$\Psi$ (ours) & 94 (\textbf{99}) & 76 (92) & \textbf{98} (\textbf{100}) & \textbf{80} (\textbf{91}) & \textbf{92} (\textbf{98}) & 75 (\textbf{100}) \\
\bottomrule
\end{tabular}
\end{table}

Table~\ref{tab:localisation} reports interchange intervention accuracy against the benchmark's published baselines. DAS is the closest method to our approach: it also fits a rotation of the residual stream at a fixed rank, but it is given an executable high-level model and a curated list of base--source pairs whose counterfactual answer is known in advance. $\Psi$ is given neither, and at its best layer it matches or exceeds DAS at all but two cells, trailing by about a point in both. 
\rev{These margins also reflect data use and training budget: the published baselines were trained on the curated pairs only and with fewer updates, and at matched data and budget $\Psi$ and DAS perform on par (Section~\ref{sec:results-supervision}).
}

On Arithmetic{\renewcommand{\thefootnote}{\fnsymbol{footnote}}\footnote[2]{Versions of \citet{mueller2025mib} disagree on the published entries. We quote the latest \href{https://arxiv.org/abs/2504.13151v2}{arXiv version}, which is the one we could reproduce from the authors' \href{https://github.com/aaronmueller/MIB}{released code}, and which differs from the proceedings only in this column. The \href{https://huggingface.co/spaces/mib-bench/leaderboard}{official leaderboard} carries no Arithmetic entry at the time of writing.}}, the depth mean puts $\Psi$ ahead on \model{gemma-2\_2b} ($\PsiArithmeticGemmaXCarryMean$ against $\DasArithmeticGemmaXCarryMean$) and close behind DAS on \model{llama-3.1\_8b} ($\PsiArithmeticLlamaXCarryMean$ against $\DasArithmeticLlamaXCarryMean$). On RAVEL (Appendix~\ref{sec:appendix-ravel}, Table~\ref{tab:ravel}), $\Psi$ \rev{trails DAS }in all measured models and variables \rev{and leads the other published methods in all but one cell}.

\begin{figure}[t]
\centering
\begin{minipage}[c]{0.40\textwidth}
\centering
\includegraphics[width=\textwidth]{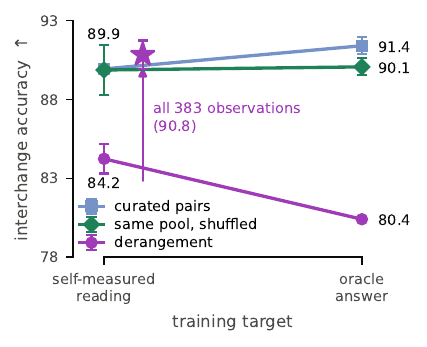}
\end{minipage}\hfill
\begin{minipage}[c]{0.56\textwidth}
\caption{\textbf{What a counterfactual dataset buys.} MCQA $X_{\mathrm{Order}}$,
\model{gemma-2\_2b}, layer $17$; mean and standard deviation over three seeds. The two ingredients of counterfactual supervision are varied independently: how base and source are paired (color) and what the intervened model is trained towards (horizontal axis). The oracle answer helps the curated arm and hurts the deranged one, so an oracle label is worth having together with the oracle's pairing. The star trains the deranged arm on all $\SupervisionEligiblePool$ eligible observations, which a derangement can use because it needs no pair list.}
\label{fig:supervision}
\end{minipage}
\end{figure}

\label{sec:results-supervision}%
\textbf{Supervision ablation.} A counterfactual dataset supplies two things: a curated pairing of each base prompt with a source, and the oracle answer that pair should produce. We focused on the MCQA task and the \model{gemma-2\_2b} model, and varied the two ingredients independently at one site (layer $17$), holding rank, optimizer and budget fixed, and restricting the unpaired arms to the $\SupervisionMatchedBases$ observations the benchmark's pair list uses as bases so that all four cells see the same prompts. Holding the pair list's own rows, its base multiplicities and its $\SupervisionMatchedUpdates$ optimizer updates, and permuting only which source accompanies which base, costs $\SupervisionCouplingGain$ points ($\SupervisionCCuratedGreedy$ against $\SupervisionShuffledSoftGreedy$) \rev{when the target is the subject's own reading, and $\SupervisionCouplingGainHard$ ($\SupervisionDCuratedHardGreedy$ against $\SupervisionShuffledHardGreedy$) when it is the oracle answer}. The coupling a counterfactual dataset supplies is therefore worth \rev{little }on its own. What it does supply is a donor pool: drawing donors from the same rows instead of from the pair list's distinct sources costs $\SupervisionPoolGain$ points ($\SupervisionDerangementSoftGreedy$). A derangement needs no pair list and can therefore draw on all $\SupervisionEligiblePool$ eligible observations, where the curated arms are confined to the $\SupervisionMatchedBases$ the pair list uses as bases; doing so reaches $\SupervisionEShippedGreedy \pm \SupervisionEShippedGreedySd$ against $\SupervisionDCuratedHardGreedy \pm \SupervisionDCuratedHardGreedySd$ for the fully supervised cell.
\rev{We confirmed this observation by training both $\Psi$ and DAS with matched data (same donor pool size) and training budget, and observed that the two methods perform on par (Appendix~\ref{sec:appendix-budget}). Additionally, we removed the correctness filter, so that no gold label enters training or checkpoint choice (Appendix~\ref{sec:appendix-filter}): this costs $\Psi$ $\RbFilterGemmaSoftDrop$ points on }\model{gemma-2\_2b} \rev{($\RbFilterGemmaFilteredSoft$ to $\RbFilterGemmaUnfilteredSoft$), and the model's distribution remains the best of the training targets we compared, on both models and with or without the filter. We conclude that the loss of pairing supervision is compensated by a larger donor pool, and that replacing the gold-label target with the model's own distribution results in better subspace identification when the model answers incorrectly (Section~\ref{sec:results-natural})}.

\rev{\textbf{Transfer across answer encodings.} The MCQA program applies unchanged to any labelling of the options, since its library is the prompt's own labels by position (Section~\ref{sec:psi}). A carrier trained on one labelling can therefore be tested on another without retraining: we froze the carriers trained at the site above and re-encoded the options of donor and recipient independently (Appendix~\ref{sec:appendix-transfer}). With lowercase labels on both sides, every learned carrier reaches $\RbTransferLowercaseMin$\%. Digit labels show an asymmetry: a digit-labelled donor moves a recipient with the original labels to the donor's position on $\RbTransferDigitDonorLow$--$\RbTransferDigitDonorHigh$\% of pairs, while a digit-labelled recipient follows the donor on only $\RbTransferDigitRecipientLow$--$\RbTransferDigitRecipientHigh$\%, although the model answers digit-labelled questions correctly on $\RbNativeDigits$\% of prompts. The carrier therefore holds the position in a form that does not depend on the donor's labels, and digit-labelled prompts do not rely on it alone to produce their answer; $\Psi$, one-hot targets and DAS at the same budget behave alike.}

\begin{figure}[t]
\centering
\includegraphics[width=0.83\textwidth]{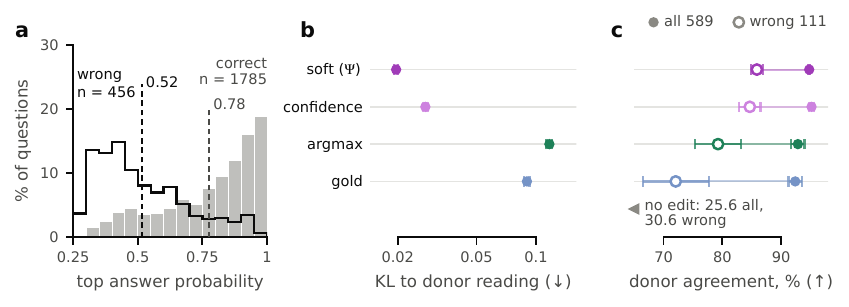}
\caption{\textbf{\rev{ARC science questions with different targets.}} \rev{Natural ARC questions, }\model{gemma-2\_2b}\rev{, layer $17$, }\textsf{\rev{last\_token}}\rev{, rank $16$; $\RbArcTrainQuestions$ unfiltered training questions.} \textbf{\rev{(a)}} \rev{Top-option probability of the unedited model on training questions it answers correctly (bars) and wrongly (line); dashed lines are the means. }\textbf{\rev{(b)}} \rev{KL from the donor's reading to the edited recipient's, on $\RbArcPairs$ held-out pairs. Rows give the training target: the model's distribution (soft, $\Psi$); its top option and probability, with the rest spread evenly (confidence); one-hot on its top option (argmax); one-hot on the gold label (gold).} \textbf{\rev{(c)}} \rev{Greedy agreement of the edited recipient with the donor's own answer, on all pairs (filled) and on the $\RbArcWrongDonors$ pairs whose donor the model answers wrongly (hollow). Markers are three seeds (mean and SD).}}
\label{fig:arc}
\end{figure}

\subsection{\rev{Supervision on natural questions}}
\label{sec:results-natural}

\rev{We then considered the natural four-choice science questions from ARC (Easy), where the four option labels are again an easy library with simple transport rules, but answering requires knowledge that no short algorithm provides, i.e.\ explicitly writing down a high-level causal model $\mathbf{H}$ is conceptually hard. We explored four different training targets: the model's distribution over the four labels ($\Psi$); a control that keeps the model's top option and its probability and spreads the remainder evenly over the other three; a one-hot label on the model's top option, which amounts to DAS with labels read from the model; and a one-hot gold label. Notably, only the last configuration makes use of gold labels, while the others rely on the model's own answers. We train on all $\RbArcTrainQuestions$ training questions, of which }\model{gemma-2\_2b} \rev{answers $\RbArcTrainAccuracy$\% correctly, at layer $17$, }\textsf{\rev{last\_token}}\rev{, rank $16$, with donors drawn from other questions, and test on $\RbArcPairs$ pairs of held-out questions and measure how closely the swapped recipient reproduces the donor's own distribution and answer (Figure~\ref{fig:arc}; Appendix~\ref{sec:appendix-arc}).}

\rev{The model's distribution is far from one-hot even when its answer is correct (Figure~\ref{fig:arc}a): on those questions, its top probability over the four labels averages $\RbArcTopCorrect$, and only $\RbArcTopConfident$\% of all questions exceed $0.9$. A one-hot target therefore differs meaningfully from the model's state even where the model is right, which is why we train on the model's distribution and on its top option as separate targets. When trained on the model's distribution ($\Psi$), the swapped recipient gives the donor's answer on $\RbArcSoftAgreement$\% of pairs, and its distribution is closer to the donor's (KL $\RbArcSoftKl$) than with one-hot model labels ($\RbArcHardKl$) or gold labels ($\RbArcGoldKl$; Figure~\ref{fig:arc}b). The source of the target matters most where the model is wrong (Figure~\ref{fig:arc}c). On the $\RbArcWrongDonors$ pairs whose donor question the model answers wrongly, the recipient takes the donor's answer on $\RbArcSoftAgreementWrong$\% of pairs with $\Psi$, $\RbArcHardAgreementWrong$\% with one-hot model labels and $\RbArcGoldAgreementWrong$\% with gold labels. We argue that a gold label on such a question describes the task, while an interchange asks about the model's state, so the subspace a gold label trains carries that state less well. The confidence control reaches $\RbArcConfidenceAgreementWrong$\%: most of the benefit comes from keeping the model's confidence, and the rest of its distribution adds some fidelity in KL. Lastly, on the benchmark's own score, which compares the output with the gold counterfactual answer on correctness-filtered pairs, the gold-trained subspace is ahead ($\RbArcGoldOfficial$ against $\RbArcSoftOfficial$), as expected for a score that rewards the correct answer.
}

\subsection{Controlling a variable without a donor}
\label{sec:results-control}

\begin{figure}[t]
\centering
\includegraphics[width=0.8\textwidth]{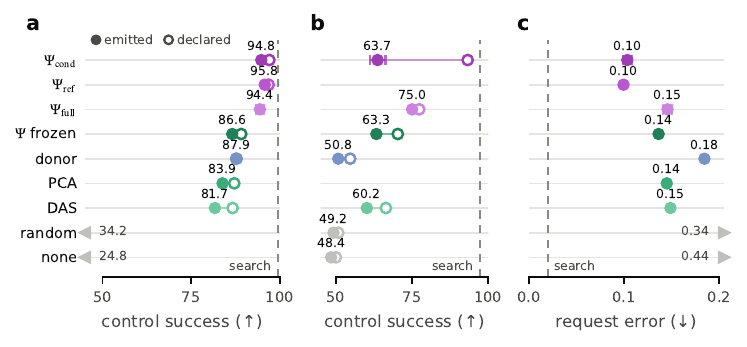}
\caption{\textbf{A declared reading can be requested.} Rows are shared across panels\rev{. The }first three train $\Psi$ jointly under each mass mode\rev{; $\Psi$ frozen (trained with $\lambda = 0$), PCA, DAS and random }attach the same controller to a frozen carrier \rev{of the same rank; }\emph{\rev{donor}} \rev{interchanges, inside the frozen $\Psi$ carrier, the training prompt whose reading is nearest the request; }\emph{\rev{none}} \rev{applies no edit}. \textbf{(a, b)} Held-out corner requests realized at MCQA $X_{\mathrm{Order}}$ layer $17$ and Arithmetic $X_{\mathrm{Carry}}$ layer $21$ (\model{gemma-2\_2b}); filled dots are greedy decoding and hollow ones the declared-library rule, on the same recipients.
\textbf{(c)} Total variation to the target on interior requests\rev{, at the MCQA site}. Seed whiskers (three seeds) mostly fall within the dots; the dashed line (\emph{search}) optimizes each edit separately.}
\label{fig:control}
\end{figure}

Correct localization does not by itself guarantee control \citep{galeone2026perfect}, and a carrier trained only to transport readings between prompts need not be writable to a reading that no prompt exhibits. We test this by training the carrier and the controller jointly, through the full objective of Eq.~\eqref{eq:full_loss}, at two sites in \model{gemma-2\_2b}: MCQA $X_{\mathrm{Order}}$ at layer $17$, where the declared library has four entries, and Arithmetic $X_{\mathrm{Carry}}$ at layer $21$, where it has two. A reading $q$ encodes a probability distribution over the declared library $\mathcal{O}$, and geometrically it is a point in the corresponding probability simplex. Its corners are the requests with an unambiguous right answer, i.e. conditions for which the model should \emph{answer as though the variable took value $v$}, while \rev{a generic request $q^\star$ is a point }in the interior of the simplex. Note that for \rev{such a request }there is typically no prompt in the data that exhibits \rev{that reading, }and hence no donor that could provide it. During training, half the requests are readings other prompts exhibit and half are uniform draws from the simplex; we evaluate on held-out corner requests, where compliant steering behavior is well-defined. We report what the subject emits under unrestricted decoding (Figure~\ref{fig:control}a,b): requests are realized $\ControlJointConditionalGreedy$\% of the time at the MCQA site and $\ArithControlJointConditionalGreedy \pm \ArithControlJointConditionalGreedySd$\% at the arithmetic one, above the donor interchange at both, which on arithmetic reaches $\ArithControlDonorGreedy$\%, at the $\ArithControlChance$\% chance floor. 
Appendix~\ref{sec:appendix-rules} \rev{compares the two decision rules}. The same controller inside a rank-matched DAS, principal or random subspace reaches $\ControlFrozenDasGreedy$\%, $\ControlFrozenPcaGreedy$\% and $\ControlFrozenRandomOneGreedy$\%. Accuracy, however, only asks which answer wins, which fails to capture the nuance of a generic request \rev{$q^\star$}. In panel (c) we thus report the total variation distance to the requested reading on the interior requests, which no donor exhibits, where $\Psi$ reaches $\ControlJointConditionalError$ against $\ControlDonorError$ for an interchange and $\ControlNoChangeError$ for no edit.

\section{Discussion}
\label{sec:discussion}
A semantic program $\mathcal{P}$ replaces the executable high-level model $\mathbf{H}$ that causal-variable methods are supervised by. In our experiments we showed that such an interface can successfully identify a low-rank causal subspace with $\Psi$ showing competitive performance against available baselines (Table~\ref{tab:localisation}), \rev{on par with the supervised alternatives at matched data and budget}. Our supervision ablation investigation (Figure~\ref{fig:supervision}) on the MCQA task corroborated these benchmark findings by revealing how the majority of the performance benefits actually stem from a net increase in the available donor pool size, with explicit counterfactual pairing only contributing modest improvements. \rev{Because the labels are read from the model, $\Psi$ also needs no ground truth: on natural questions, the resulting subspace tracks the model's own answers, wrong ones included, better than one trained on gold labels (Figure~\ref{fig:arc}), which opens interchange-based localization to tasks where no labels exist}. We showed that joint localization and control training yields better control performance (Figure~\ref{fig:control}), both in terms of explicit control successes and request errors, than than a controller built on top of a frozen localization-only carrier, which is in line with previous observations that explicitly differentiate the two concepts \citep{galeone2026perfect}.

\textbf{Limitations.}
\label{sec:discussion-limitations}
First, the semantic programs in this paper were easy to write. Their tasks have small, explicit answer sets, and for a more abstract variable it may be hard to name a library that is both semantically clean and broad enough to hold much of the model's probability. \rev{How much a library saves also depends on the variable: for the carry in addition, writing it requires solving the task. Moreover, targets read from the model are only as good as the model's distribution over the library: on RAVEL, where this distribution is diffuse, $\Psi$ trails DAS (Appendix~\ref{sec:appendix-ravel})}. Another limitation is that interchange accuracy by itself is not a guarantee of identification: a subspace that supports the declared transport $\bm{\tau}$ need not be unique; moreover, a subspace can move the output through a pathway that has nothing to do with the variable it is supposed to name \citep{makelov2024subspace}. \rev{Finally, }gold answers are used to filter the training pool \rev{of Table~\ref{tab:localisation}}, as they are for the published baselines \citep{mueller2025mib}\rev{, although removing the filter costs }$\Psi$ \rev{less than a point (Section~\ref{sec:results-supervision})}.

\subsection*{AI use statement}

In this work we have used generative AI tools to refine hypotheses, provide feedback on experiments and assist the experiment lifecycle (e.g.\ submitting cluster jobs, writing scripts to collect results). We have not used AI tools to design the research, write the core $\Psi$ algorithm, support qualitative and thematic data analysis, interpret results, generate synthetic datasets or assist with translation; the remaining tasks (e.g.\ theoretical proofs) are not applicable to this work.
Additionally, we used generative AI tools to create or modify scientific figures, draft sections of the manuscript, write supporting code (e.g.\ tests, evaluation scripts), and summarize existing literature and identify gaps. We have reviewed all AI-assisted work, inspecting the code and writing and editing the full manuscript ourselves. We take responsibility for the final content of this work, including text, claims or artifacts produced with the aid of generative AI.

\subsection*{Reproducibility statement}
Appendix~\ref{sec:appendix-programs} gives the complete semantic program for
every task, which together with the objective of
Section~\ref{sec:objective} and the public MIB dataset \citep{mueller2025mib} is everything an implementation needs to reproduce our results.

\bibliographystyle{iclr2027_conference}
\bibliography{iclr2027_conference}

\newpage

\appendix
\section{Appendix}
\label{sec:appendix}
\subsection{Additional related work}
\label{sec:appendix-related}

\paragraph{The distributed alignment lineage.} DAS learns a rotation whose coordinate block aligns with a high-level variable \citep{geiger2024finding}, Boundless DAS learns the subspace boundary as well so the rank need not be fixed by search \citep{wu2023boundless}, MDAS optimizes attribute isolation alongside
the alignment on RAVEL \citep{huang2024ravel}, and HyperDAS replaces the brute-force search over feature locations with a hypernetwork
\citep{sun2025hyperdas}. All four take their training target from an executable high-level model, which is the input a semantic program replaces. HyperDAS in particular automates where to look; the declaration changes what the investigator has to write down, and the two are independent.

\paragraph{Writing a behavior into activations.} Representation engineering \citep{zou2023representation} and activation addition \citep{turner2023steering, rimsky2024steering} add to the residual stream a direction fitted from contrastive prompts, and inference-time intervention does the same along a truth-related direction \citep{li2023inference}. Recent work learns the write operator instead of fitting it. CDAS matches the intervened output distribution against the model's own distribution on concept-eliciting prompts and clamps a rank-1 coordinate at inference \citep{bao2026faithful}; a single causally identified direction acts as a context-sensitivity knob across a model family \citep{minder2025controllable}; LinEAS fits per-layer affine maps against a distributional loss over activation populations \citep{rodriguez2025lineas}; a probe-free low-rank map pushes an activation towards a manifold of desirable content \citep{jiang2025probe}; and UniSteer conditions a flow in activation space on a natural-language description of the wanted behavior
\citep{shi2026unisteer}. Each of weak supervision, model-relative targets, donor-free inference and amortization over a space of requests is therefore established on its own. What a semantic program adds is a request that is a distribution over a declared answer set indexed by a causal variable's values, issued to a subspace that interchange interventions have independently validated.

The write operator itself is LoReFT's, i.e. project into a low-rank subspace, compute a target there, add the residual back \citep{wu2024reft}. Ours takes the requested reading as an argument, so one training run covers the whole simplex rather than one behavior.

\paragraph{Localization and control are different measurements.} Detection directions and steering directions are close to orthogonal \citep{galeone2026perfect}, and the methods that win at concept detection and at steering are not the same ones \citep{wu2025axbench}. The geometry says why: the directions that read a concept off an activation and the directions that change it are related by the causal inner product and not by the Euclidean one \citep{park2024linear}. Steering vectors are in addition variable across inputs and brittle to reasonable changes of prompt \citep{tan2024analysing}, and a subspace intervention can move the output through a dormant parallel pathway while the feature it names does nothing \citep{makelov2024subspace}. Intervention success as an evaluation criterion is itself established \citep{bhalla2025unifying}; we use it as a second measurement on one identified carrier rather than as a faithfulness criterion in its own right.

\subsection{Supervision accounting}
\label{sec:appendix-supervision}

\begin{table}[ht]
\centering
\footnotesize
\setlength{\tabcolsep}{6pt}
\caption{Supervision accounting. Each column is an input the investigator must provide before training begins, and the entries say \emph{what} that input is: the methods differ in kind, not only in quantity. The DAS lineage \citep{geiger2024finding, wu2023boundless, huang2024ravel} shares one row because it asks for one thing. PSI replaces the algorithmic specification with a declaration in the answer space and the oracle counterfactual with a measurement; gold answers still enter, as the correctness filter on the training pool that the published baselines also apply. Remaining rows: HyperDAS \citep{sun2025hyperdas}, CDAS \citep{bao2026faithful}, LoReFT \citep{wu2024reft}.}
\label{tab:supervision}
\begin{tabular}{@{}lcccc@{}}
\\
\toprule
Method & High-level model & Intervention target & Pairing & Gold labels \\
\midrule
DAS, Boundless DAS, MDAS & algorithm          & oracle answer     & curated       & target, filter \\
HyperDAS                 & algorithm, text    & oracle answer     & curated       & target, filter \\
CDAS                     & concept prompts    & own distribution  & contrast sets & set membership \\
LoReFT                   & none               & task label        & none          & training target \\
\midrule
$\Psi$ (ours)            & library, transport & own reading       & derangement   & filter \\
\bottomrule
\end{tabular}
\end{table}

The row order runs from the most to the least that a method asks of the investigator before training begins. Every entry is an input, so a shorter row is a cheaper method to apply, and Section~\ref{sec:results-supervision} measures what the entries in the DAS row are worth in accuracy.

\subsection{The semantic program for each task}
\label{sec:appendix-programs}

Everything the investigator writes for a task is below. Each program is an outcome library $\mathcal{O}$ and a transport rule $\mathcal{T}$, fixed before training and shared by every run on that task. Where a rule needs the source's reading and the source prompt does not supply one, we also give the query that measures it.

\paragraph{MCQA and ARC (Easy).} Both are multiple-choice, and both variables use the full-change rule $\bm{\tau} = q(\mathbf{s})$. They differ in what the library names and in how wide it is. For $X_{\mathrm{Order}}$, the position of the correct option, the library is the prompt's own choice labels in the order that prompt lists them, each with its leading space, indexed by option position; an interchange that moves the variable moves the model from one label to another. For $O_{\mathrm{Answer}}$, the answer symbol, the library is the twenty-six leading-space letters $(\texttt{\char32A}, \dots, \texttt{\char32Z})$, identical for every prompt. The two variables therefore do not share a library: one has four entries drawn from the prompt, the other twenty-six fixed ones, which is what lets a single benchmark task carry two variables at different widths.

\paragraph{Arithmetic (+).} The variable is the carry out of the units digits, so the library has two entries: the sum with the carry propagated and the sum without it. For $\texttt{47 + 28 =}$ these are $\texttt{75}$ and $\texttt{65}$. Both are computed from the operands the prompt states and not from the answer, but writing that library does take knowing what a carry does to a sum. The program supplies the mapping from this variable's values to the strings they produce, and nothing else about how the subject computes them. The rule is again the full change. The carry is a natural variable to declare: it is the intermediate the benchmark itself targets \citep{mueller2025mib}, transformers trained on addition compute it explicitly \citep{quirke2024understanding}, and causal mediation analysis has localized where language models route the operands of such sums \citep{stolfo2023mechanistic}.

\paragraph{RAVEL.} The library is built once from the benchmark's entity file: every value any city takes on any of the three attributes, reduced to its first token under the subject's tokenizer and deduplicated, so a single library of country, continent and language names serves all three variables. Entries are therefore first tokens rather than complete attribute names, and two values sharing a first token are one entry. The rule is gated, $\bm{\tau} = \gamma (\mathbf{b})\, q(\mathbf{s}) + (1 - \gamma (\mathbf{b}))\, q(\mathbf{b})$, with $\gamma (\mathbf{b}) = 1$ when the base prompt queries the attribute being intervened on and $\gamma (\mathbf{b}) = 0$ otherwise. That single bit is the whole of what RAVEL's isolation requirement asks for: change the answer when the question is about the attribute you moved, leave it alone when it is not.

RAVEL is also the one task where the source's reading cannot be taken from the source prompt. It has two prompt families: JSON-style continuations such as \texttt{\{"city": "Nagoya", "language": "}, which name an attribute, and free text such as \texttt{Spartak Trnava won 3--1 on aggregate}, which names none. The program therefore declares the query that measures the source's value,

\begin{center}
\texttt{Q: What is the \{attribute\} of \{entity\}?}\\
\texttt{A:}
\end{center}

evaluated on the frozen subject with \texttt{\{attribute\}} the one being intervened on. This is the declared query of Eq.~\eqref{eq:transport_rule}: one extra forward pass per source observation, and the only place a program in this paper reaches beyond the prompts it was given.

\paragraph{IOI.} We consider the IOI task an extension rather than an instance of the interface, and we separate it because its transport rule is not measured from the subject, due to the particular implementation of this task in the MIB benchmark itself. The benchmark scores the task as a signed logit margin between the indirect object and the subject name, and supplies a linear model of that margin, $\mathrm{bias} + \beta_{\mathrm{tok}} s_{\mathrm{tok}} + \beta_{\mathrm{pos}} s_{\mathrm{pos}}$, with fixed coefficients and each signal $+1$ when that variable still matches the base and $-1$ when it does not. We build the transport table by evaluating that model under the isolated intervention and passing the resulting margin through a sigmoid, which gives the probability the indirect object should carry; the library is the two names. The numbers in an IOI target therefore come from the benchmark's high-level model and from the gold identity of the base's indirect object, not from the frozen subject. Every claim in this paper about targets measured from the subject covers MCQA, ARC, arithmetic and RAVEL; it does not cover IOI, and Appendix~\ref{sec:appendix-ravel} reports that task on those terms.

\paragraph{Unused supervision.} With the exception of IOI above, no
program names a correct answer, and none names which library entry a given prompt should elicit. Gold answers enter the other four tasks outside the program, as the filter that keeps only observations the subject already answers correctly, and through the benchmark's own choice of cut position, which is located by the correct answer symbol. Both are conventions the published baselines also follow\rev{, and the experiments of Section~\ref{sec:results-natural} use neither}.

\subsection{Two decision rules}
\label{sec:appendix-rules}

The benchmark's evaluator decodes from the intervened model over the whole vocabulary and asks whether the expected answer appears. Our training objective works instead on the declared library, so we also audit with the rule the objective optimizes: restricted to the library, does the intervened model prefer the entry the transport rule asked for? Comparing the two tells us whether a score reflects the variable moving or the decode landing elsewhere.

The two agree closely wherever the declared answers are single tokens. They come apart on arithmetic, where an answer spans three tokens: at layer $\PsiArithmeticGemmaXCarryLayer$ the carrier scores $\PsiArithmeticGemmaXCarryBest$ under the benchmark's rule and $\PsiArithmeticGemmaXCarryDeclaredBest$ under ours. The carry does move, and the rollout that follows it often lands on a number outside the declared pair. Every localization number in the main text uses the benchmark's rule; Section~\ref{sec:results-control} reports both.

\subsection{The mass term}
\label{sec:appendix-mass}

Equation~\eqref{eq:decomposition} splits the training signal into the shape of the reading and the probability mass the library holds, and Section~\ref{sec:objective} leaves open what to ask of the second term. We measured all three options at one site over three seeds to choose between them. Dropping the term (\textsc{conditional}), aiming it at the mass the unintervened prompt already shows (\textsc{reference}), and maximizing it (\textsc{full}) give $\MassMcqaConditionalGreedy \pm \MassMcqaConditionalGreedySd$, $\MassMcqaReferenceGreedy \pm \MassMcqaReferenceGreedySd$ and $\MassMcqaFullGreedy \pm \MassMcqaFullGreedySd$ on MCQA $X_{\mathrm{Order}}$. \textsc{full} is the only mode that pays to keep probability on the declared answers, which is what should stop an undeclared token winning a decode, and it scores lowest here, so we use \textsc{conditional} for every task except Arithmetic. There the answers span several tokens, and preserving the mass on the digits benefits greedy decoding: at the \model{gemma-2\_2b} $X_{\mathrm{Carry}}$ site (layer $\MassArithmeticLayer$, three seeds) \textsc{full} reaches $\MassArithmeticFullGreedy \pm \MassArithmeticFullGreedySd$ against $\MassArithmeticConditionalGreedy \pm \MassArithmeticConditionalGreedySd$ for \textsc{conditional} under the benchmark's rule, even though it trails under the declared one ($\MassArithmeticFullDeclared$ against $\MassArithmeticConditionalDeclared$). The control experiments show the same ordering under greedy decoding (Figure~\ref{fig:control}b and below), so the Arithmetic localization runs use \textsc{full}.

The same choice is equally important for control, where the two terms of the decomposition pull against each other. At the arithmetic site $\Psi_{\mathrm{full}}$ keeps $\ArithControlJointFullMass$ of the declared mass after the edit against $\ArithControlJointConditionalMass$ for $\Psi_{\mathrm{cond}}$, and pays $\ArithControlMassTrade$ points of control success under the declared rule for it ($\ArithControlJointFullDeclared$\% against $\ArithControlJointConditionalDeclared \pm \ArithControlJointConditionalDeclaredSd$\%), and buys it back under free decoding, where the order reverses ($\ArithControlJointFullGreedy$\% against $\ArithControlJointConditionalGreedy \pm \ArithControlJointConditionalGreedySd$\%): the mass \textsc{conditional} gives up is mass the undeclared tail takes. Asking the model to keep talking about the declared answers and asking it to change which one it gives are different requests, and the decomposition is where a program states which of the two it wants.

\subsection{What the control term costs}
\label{sec:appendix-joint}
\label{sec:appendix-cost}

\paragraph{Carrier cost.} The two terms of Eq.~\eqref{eq:full_loss} share the subspace $A$, so it is reasonable to expect them to compete: a subspace shaped to be writable need not be the best carrier under transplantation. We measured the effect at the MCQA $X_{\mathrm{Order}}$ site of \model{gemma-2\_2b}, training three seeds either for localization alone ($\lambda = 0$ in Eq.~\eqref{eq:full_loss}) or jointly ($\lambda = 1$). Seed for seed, turning the control term on lowers interchange accuracy by $\ControlTermCost$ points and raises control success by $\ControlJointOverFrozen$. The cost is small but consistent, and it is why the benchmark evaluations of Section~\ref{sec:results-localisation} use $\lambda = 0$ while the control experiments use $\lambda = 1$.

The joint carrier is also trained for longer than the frozen ones it is compared against, so we checked whether its advantage is the training budget. Attaching the same controller to donor-only carriers trained for $8$, $32$ and $64$ epochs gives (under the declared rule) $\BudgetFrozenEight$\%, $\BudgetFrozenThreeTwo$\% and $\BudgetFrozenSixFour$\% of requests realized, against $\BudgetJoint$\% for the jointly trained carrier on the same recipients \rev{(all four with seed $11$, whereas Figure~\ref{fig:control} reports the three-seed mean)}. Four to eight times the carrier budget is worth under a point and then saturates, so we attribute the gap to full joint training.

\begin{figure}[t]
\centering
\includegraphics[width=\textwidth]{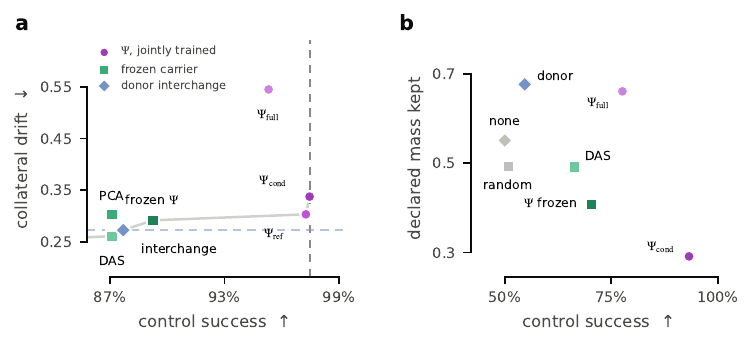}
\caption{Cost against success for the arms of Figure~\ref{fig:control}, seed $11$, the same held-out recipients and the same corner requests, so the two axes are comparable point by point.
\textbf{(a)} MCQA $X_{\mathrm{Order}}$: collateral drift is the total-variation distance between the undeclared tails of the output distribution before and after the edit, each renormalized, and the dashed line marks what a donor interchange costs at the same site.
\textbf{(b)} Arithmetic answers span several tokens, so the undeclared tail is ill defined there and the cost axis is the probability mass the declared library retains.}
\label{fig:control-cost}
\end{figure}

\paragraph{Collateral drift.} The control term also competes with the rest of the model's behavior and altering the model's preferences within the declared library $\mathcal{O}$ might come at the expense of the overall probability mass outside the library, i.e. on $\mathcal{O}_\perp$. Figure~\ref{fig:control-cost} plots the measured controlled accuracies for various methods against a measure of collateral control cost, revealing a Pareto-like tradeoff between control efficacy and intervention cost.

We quantified the collateral cost differently between the MCQA task (Figure~\ref{fig:control-cost}a) and the arithmetic task (Figure~\ref{fig:control-cost}b), given that what can be measured differs by site. MCQA's declared outcomes are disjoint single tokens, so the undeclared tail is well defined and drift can be read over the whole vocabulary; we thus quantified the cost as the total variation between the tail of the distribution before the edit and the tail of the distribution after the edit. Arithmetic's answers are three tokens long, and there the available axis is how much probability the declared library still holds after the edit.

\subsection{\rev{Correctness filter}}
\label{sec:appendix-filter}

\begin{table}[ht]
\centering
\footnotesize
\setlength{\tabcolsep}{5pt}
\caption{\rev{\textbf{Correctness filter and training target.} MCQA $X_{\mathrm{Order}}$, layer $17$, \textsf{last\_token}, rank $16$; $\RbBudgetUpdates$ updates, final checkpoint, seeds $11$--$13$ (mean $\pm$ SD). Filtered: training prompts the model answers correctly ($\RbFilterGemmaPool$/$\RbFilterGemmaPoolAll$ for \model{gemma-2\_2b}, $\RbFilterQwenPool$/$\RbFilterQwenPoolAll$ for \model{qwen-2.5\_0.5b}); unfiltered: all prompts. On the filtered pool the model's argmax is the gold label, so the two arms are the same runs. Official: the benchmark's evaluator ($\RbOfficialPairs$ eligible pairs). Agreement: edited argmax equals the donor model's argmax; KL: donor reading to edited reading over the four options; both on the same $150$ test pairs for every arm.}}
\label{tab:rebuttal-filter}
\begin{tabular}{@{}llccc@{}}
\\
\toprule
Pool & Target & Official (\%) $\uparrow$ & Agreement (\%) $\uparrow$ & KL $\downarrow$ \\
\midrule
\multicolumn{5}{@{}l}{\emph{\model{gemma-2\_2b}}} \\
filtered & soft reading ($\Psi$) & $\mathbf{90.24} \pm 0.52$ & $\mathbf{82.22} \pm 0.38$ & $0.103 \pm 0.001$ \\
filtered & model argmax = gold & $88.55 \pm 0.49$ & $79.56 \pm 1.02$ & $0.277 \pm 0.009$ \\
unfiltered & soft reading ($\Psi$) & $89.34 \pm 0.52$ & $\mathbf{82.22} \pm 1.39$ & $\mathbf{0.102} \pm 0.004$ \\
unfiltered & model argmax & $86.44 \pm 0.51$ & $79.56 \pm 0.38$ & $0.149 \pm 0.009$ \\
unfiltered & gold label & $87.59 \pm 1.43$ & $78.22 \pm 1.02$ & $0.280 \pm 0.006$ \\
\midrule
\multicolumn{5}{@{}l}{\emph{\model{qwen-2.5\_0.5b}}} \\
filtered & soft reading ($\Psi$) & $\mathbf{99.57} \pm 0.74$ & $82.00 \pm 0.00$ & $0.194 \pm 0.002$ \\
filtered & model argmax = gold & $96.65 \pm 0.69$ & $80.00 \pm 0.00$ & $1.030 \pm 0.019$ \\
unfiltered & soft reading ($\Psi$) & $97.78 \pm 2.96$ & $82.44 \pm 1.02$ & $\mathbf{0.180} \pm 0.002$ \\
unfiltered & model argmax & $96.99 \pm 2.82$ & $\mathbf{82.89} \pm 0.77$ & $0.298 \pm 0.007$ \\
unfiltered & gold label & $97.08 \pm 1.60$ & $79.78 \pm 0.77$ & $1.141 \pm 0.044$ \\
\bottomrule
\end{tabular}
\end{table}

\rev{The training pools of the main experiments keep only prompts the model answers correctly. Table~\ref{tab:rebuttal-filter} trains the same carrier on that filtered pool and on all prompts, with three training targets, a fixed budget of $\RbBudgetUpdates$ updates and the final checkpoint, so no gold label enters through checkpoint selection. Without the filter, the soft-reading carrier scores $\RbFilterGemmaSoftDrop$ points lower on }\model{gemma-2\_2b} \rev{and $\RbFilterQwenSoftDrop$ points lower on }\model{qwen-2.5\_0.5b} \rev{under the official evaluator.
}

\subsection{\rev{Matched training budget}}
\label{sec:appendix-budget}

\begin{table}[ht]
\centering
\footnotesize
\setlength{\tabcolsep}{5pt}
\caption{\rev{\textbf{Matched training budget.} MCQA $X_{\mathrm{Order}}$, \model{gemma-2\_2b}, layer $17$, \textsf{last\_token}, rank $16$, seeds $11$--$13$. Every arm takes $\RbBudgetUpdates$ optimizer updates over $\RbBudgetExamples$ training examples with the same batch sizes, and is scored at its final checkpoint. The $\Psi$ arms use the $\RbBudgetPsiObservations$ correctly answered training prompts with deranged donors; DAS uses the benchmark's curated pairs ($\RbBudgetDasPairs$ after filtering). Official: the benchmark's evaluator on $\RbOfficialPairs$ pairs. KL: donor reading to edited reading on $150$ test pairs; not computed for DAS.}}
\label{tab:rebuttal-budget}
\begin{tabular}{@{}llcc@{}}
\\
\toprule
Target & Mass mode & Official (\%) $\uparrow$ & KL $\downarrow$ \\
\midrule
soft reading ($\Psi$) & \textsc{conditional} & $90.24 \pm 0.52$ & $0.103 \pm 0.001$ \\
soft reading ($\Psi$) & \textsc{reference} & $\mathbf{92.00} \pm 1.40$ & $\mathbf{0.102} \pm 0.001$ \\
soft reading ($\Psi$) & \textsc{full} & $89.03 \pm 0.00$ & $0.110 \pm 0.003$ \\
model argmax & \textsc{conditional} & $88.55 \pm 0.49$ & $0.277 \pm 0.009$ \\
model argmax & \textsc{reference} & $89.71 \pm 0.95$ & $0.391 \pm 0.007$ \\
model argmax & \textsc{full} & $87.64 \pm 1.07$ & $0.195 \pm 0.010$ \\
\midrule
DAS (official implementation) & -- & $91.70 \pm 0.93$ & -- \\
\bottomrule
\end{tabular}
\end{table}

\rev{Table~\ref{tab:rebuttal-budget} compares $\Psi$ with the benchmark's DAS implementation retrained at the same budget: the same number of optimizer updates, training examples and batch sizes, and the final checkpoint. The $\Psi$ rows cross the three mass modes of Section~\ref{sec:objective} with soft and argmax targets. Differences that remain: DAS keeps its own optimizer and full-vocabulary objective, and trains on the benchmark's curated pairs.
}

\subsection{\rev{Official score by family}}
\label{sec:appendix-families}

\begin{table}[ht]
\centering
\footnotesize
\setlength{\tabcolsep}{5pt}
\caption{\rev{\textbf{Official score by counterfactual family.} Top: the arms of Table~\ref{tab:rebuttal-budget} (\textsc{conditional} for $\Psi$ and argmax), official evaluator, by family: only the answer's position differs between base and source ($n=\RbOfficialPairsPosition$), only the option symbols differ ($n=\RbOfficialPairsSymbol$), both differ ($n=\RbOfficialPairsBoth$); macro is their unweighted mean. Bottom: test pairs of the \emph{both} family whose donor answer symbol differs from the recipient's symbol at the donor's answer position ($n=\RbLiteralBothPairs$); share of greedy outputs equal to the recipient's symbol at the donor's position, and equal to the donor's own symbol. Greedy output: first generated field, punctuation stripped, case-sensitive. Learned arms: seeds $11$--$13$; controls: single runs.}}
\label{tab:rebuttal-families}
\begin{tabular}{@{}lcccc@{}}
\toprule
Training target & Position (\%) & Symbol (\%) & Both (\%) & Macro (\%) \\
\midrule
soft reading ($\Psi$) & $100.0 \pm 0.0$ & $\mathbf{91.4} \pm 0.0$ & $79.3 \pm 1.6$ & $90.24 \pm 0.52$ \\
model argmax & $100.0 \pm 0.0$ & $81.0 \pm 1.6$ & $84.7 \pm 3.1$ & $88.55 \pm 0.49$ \\
DAS & $100.0 \pm 0.0$ & $89.5 \pm 1.6$ & $\mathbf{85.6} \pm 1.6$ & $\mathbf{91.70} \pm 0.93$ \\
\bottomrule
\end{tabular}
\par\bigskip
\begin{tabular}{@{}lcc@{}}
\toprule
 & Recipient's symbol at & Donor's \\
Arm & donor's position (\%) & symbol (\%) \\
\midrule
soft reading ($\Psi$) & $66.7 \pm 3.5$ & $9.1 \pm 0.0$ \\
model argmax & $72.7 \pm 2.3$ & $6.8 \pm 0.0$ \\
DAS & $\mathbf{77.3} \pm 2.3$ & $5.3 \pm 1.3$ \\
full residual copy & $45.5$ & $13.6$ \\
no edit & $0.0$ & $6.8$ \\
\bottomrule
\end{tabular}
\end{table}

\rev{Table~\ref{tab:rebuttal-families} splits the official score of the matched-budget arms into the benchmark's three counterfactual families. Its lower part takes the family in which both the answer's position and the option symbols differ between base and source, and counts how often the edited output is the recipient's symbol at the donor's position and how often it is the donor's own symbol.
}

\subsection{\rev{Transfer across answer encodings}}
\label{sec:appendix-transfer}

\begin{figure}[ht]
\centering
\includegraphics[width=\textwidth]{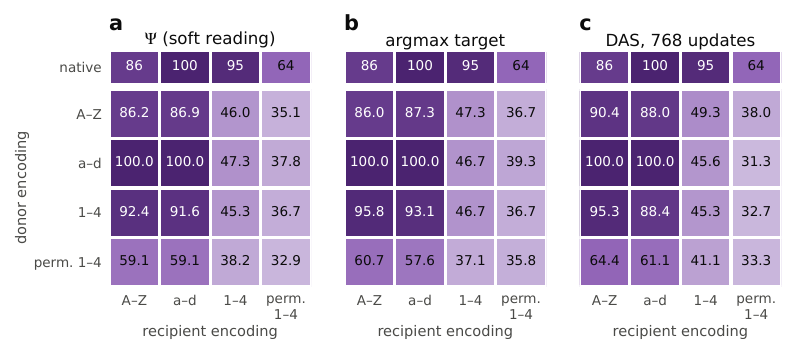}
\caption{\textbf{\rev{Transfer across answer encodings.}} \rev{MCQA $X_{\mathrm{Order}}$, }\model{gemma-2\_2b}\rev{, layer $17$, }\textsf{\rev{last\_token}}\rev{, rank $16$; carriers trained with the original labels and frozen; seeds $11$--$13$, final checkpoint of $\RbBudgetUpdates$ updates, mean over seeds. Rows: donor encoding; columns: recipient encoding (A--Z: the original uppercase labels; a--d; 1--4; perm.\ 1--4: digits permuted per prompt). Each cell is the percentage of the $\RbTransferPairs$ test pairs on which the edited recipient answers with the donor's option position written in the recipient's labels. Scoring rule: greedy, first generated field with punctuation stripped, exact case-sensitive label, unweighted mean over the three counterfactual families. The strip above each map is the unedited model's greedy accuracy on $\RbNativePrompts$ prompts under each recipient encoding. One colour scale, $0$--$100$, for all panels. }\textbf{\rev{(a)}} \rev{$\Psi$, soft reading; }\textbf{\rev{(b)}} \rev{argmax target; }\textbf{\rev{(c)}} \rev{the benchmark's DAS implementation at the same budget.}}
\label{fig:encoding-transfer}
\end{figure}

\begin{table}[ht]
\centering
\footnotesize
\setlength{\tabcolsep}{5pt}
\caption{\rev{\textbf{Controls for Figure~\ref{fig:encoding-transfer}.} Same pairs and scoring rule as the figure (greedy, family macro, all $\RbTransferPairs$ pairs); single runs.}}
\label{tab:rebuttal-transfer}
\begin{tabular}{@{}llcccc@{}}
\toprule
 & & \multicolumn{4}{c}{Recipient encoding} \\
\cmidrule(l){3-6}
Control & Donor encoding & original & lowercase & digits & perm.\ digits \\
\midrule
random subspace & original & 33.3 & 33.3 & 32.0 & 32.0 \\
 & lowercase & 33.3 & 33.3 & 31.3 & 32.7 \\
 & digits & 33.3 & 33.3 & 32.0 & 32.7 \\
 & perm.\ digits & 33.3 & 33.3 & 32.0 & 32.7 \\
\midrule
no edit & original & 33.3 & 33.3 & 32.0 & 32.7 \\
 & lowercase & 33.3 & 33.3 & 32.0 & 32.7 \\
 & digits & 33.3 & 33.3 & 32.0 & 32.7 \\
 & perm.\ digits & 33.3 & 33.3 & 32.0 & 32.7 \\
\midrule
full residual copy & original & 75.3 & 62.7 & 7.3 & 14.7 \\
 & lowercase & 100.0 & 100.0 & 4.7 & 8.7 \\
 & digits & 86.0 & 79.3 & 49.3 & 28.0 \\
 & perm.\ digits & 48.7 & 44.7 & 36.7 & 34.7 \\
\bottomrule
\end{tabular}
\end{table}

\rev{We evaluated frozen carriers, trained with the original option labels, after re-encoding the option labels of donor and recipient independently: lowercase letters, the digits $1$--$4$, and the same digits permuted per prompt. No carrier was retrained, and the expected answer is the donor's option position written in the recipient's labels. Figure~\ref{fig:encoding-transfer} gives the learned carriers and Table~\ref{tab:rebuttal-transfer} the controls on the same pairs. In a separate test with the donor's own labels lowercased and an uppercase recipient, no learned carrier outputs the donor's lowercase symbol ($\RbLiteralCasefoldLearnedCopy$\% of $\RbLiteralCasefoldPairs$ pairs); copying the full residual vector outputs the donor's letter on $\RbLiteralDigitsFullcopyCopy$\% of pairs with an original donor and a digit recipient.
}

\subsection{\rev{Natural ARC}}
\label{sec:appendix-arc}

\begin{table}[ht]
\centering
\footnotesize
\setlength{\tabcolsep}{3.5pt}
\caption{\rev{\textbf{Natural ARC by training target.} \model{gemma-2\_2b}, layer $17$, \textsf{last\_token}, rank $16$; $\RbArcTrainQuestions$ unfiltered training questions, $\RbArcUpdates$ updates, final checkpoint, seeds $11$--$13$. KL: donor's native reading to the edited reading over the four options, on $\RbArcPairs$ held-out pairs (seed mean); fitted: after one temperature per checkpoint, fitted on $\RbArcValidationPairs$ separate validation pairs (range over seeds in the next column). Agreement: greedy answer of the edited recipient equals the donor's own preferred answer, on all pairs and on the $\RbArcWrongDonors$ pairs whose donor the model answers wrongly. Official: the benchmark's evaluator on its correctness-filtered counterfactual pairs, a different population from the natural pairs.}}
\label{tab:rebuttal-arc}
\begin{tabular}{@{}lcccccc@{}}
\\
\toprule
 & \multicolumn{2}{c}{KL $\downarrow$} & & \multicolumn{2}{c}{Agreement (\%) $\uparrow$} & \\
\cmidrule(lr){2-3}\cmidrule(lr){5-6}
Target & raw & fitted & Temperature & all & wrong donor & Official (\%) \\
\midrule
soft ($\Psi$) & $\mathbf{0.020}$ & $\mathbf{0.019}$ & 1.01--1.02 & $94.8 \pm 0.2$ & $\mathbf{85.9} \pm 1.0$ & $70.03 \pm 1.21$ \\
confidence & $0.028$ & $0.028$ & 0.99--1.00 & $\mathbf{95.2} \pm 0.4$ & $84.7 \pm 1.8$ & $69.95 \pm 0.69$ \\
argmax & $0.117$ & $0.063$ & 1.56--1.58 & $92.9 \pm 1.1$ & $79.3 \pm 3.9$ & $68.27 \pm 0.81$ \\
gold & $0.090$ & $0.058$ & 1.37--1.42 & $92.4 \pm 1.1$ & $72.1 \pm 5.6$ & $\mathbf{71.45} \pm 1.36$ \\
\midrule
no edit & $1.548$ & -- & -- & $25.6$ & $30.6$ & -- \\
\bottomrule
\end{tabular}
\end{table}

\rev{The experiment of Section~\ref{sec:results-natural} uses the ARC (Easy) four-choice questions as given, on }\model{gemma-2\_2b} \rev{at layer $17$, }\textsf{\rev{last\_token}}\rev{, rank $16$. The $\RbArcTrainQuestions$ training questions are not filtered for correctness (the model answers $\RbArcTrainAccuracy$\% of them correctly), donors are other training questions, and every arm takes $\RbArcUpdates$ updates and keeps the final checkpoint. Gold labels enter the gold arm only; for the other three they are removed from the training cache and play no part in filtering, targets or checkpoint choice.
}

\rev{The test set excludes the $\RbArcExcluded$ test questions whose text also occurs among the training questions; the remaining $\RbArcTestQuestions$ distinct questions form $\RbArcPairs$ disjoint donor--recipient pairs, $\RbArcWrongDonors$ of them with a donor the model answers wrongly. Temperatures were fitted per checkpoint on $\RbArcValidationPairs$ pairs of $\RbArcValidationQuestions$ further validation questions, excluding one whose text overlaps the training or test questions, and applied unchanged to the test pairs. Table~\ref{tab:rebuttal-arc} also gives the official MIB score of each arm, which is computed on the benchmark's correctness-filtered counterfactual pairs, a different population from the natural pairs.
}

\subsection{RAVEL and IOI}
\label{sec:appendix-ravel}

\begin{table}[t]
\centering
\footnotesize
\setlength{\tabcolsep}{3.5pt}
\caption{\textbf{RAVEL and IOI.} Top, RAVEL: conventions as in Table~\ref{tab:localisation}, interchange accuracy with the best layer in parentheses, higher better. Bottom, IOI on \model{gpt2-small}, whose site is a fixed set of attention heads (three for $S_{\mathrm{Pos}}$, one for $S_{\mathrm{Tok}}$) rather than a swept layer. The benchmark scores it as a mean-squared error between the causal model's logit difference and the subject's, so lower is better there; we add the interchange accuracy of the same runs, which the publication does not report (for Full Vector, the whole-vector interchange at the same heads, from our own audit), and our own run of DAS through the same evaluator, because the published $S_{\mathrm{Tok}}$ entry is one we could not reproduce. Accuracy is not bolded where doing nothing already scores highest. A dash marks a number the source does not provide.}
\label{tab:ravel}
\begin{tabular}{@{}l ccc ccc@{}}
\\
\toprule
 & \multicolumn{6}{c}{RAVEL} \\
\cmidrule(l){2-7}
 & \multicolumn{3}{c}{\model{gemma-2\_2b}} & \multicolumn{3}{c}{\model{llama-3.1\_8b}} \\
\cmidrule(lr){2-4}\cmidrule(l){5-7}
Method & $A_{\mathrm{Cont}}$ & $A_{\mathrm{Ctry}}$ & $A_{\mathrm{Lang}}$ & $A_{\mathrm{Cont}}$ & $A_{\mathrm{Ctry}}$ & $A_{\mathrm{Lang}}$ \\
\midrule
\nul{Full Vector} & \nul{48 (62)} & \nul{49 (57)} & \nul{45 (56)} & \nul{53 (62)} & \nul{47 (53)} & \nul{47 (57)} \\
\nul{Random} & \rev{\nul{61 (62)}} & \rev{\nul{50 (51)}} & \rev{\nul{55 (55)}} & \rev{\nul{62 (62)}} & \rev{\nul{51 (51)}} & \rev{\nul{56 (57)}} \\
DBM & 66 (71) & 53 (65) & 54 (58) & 68 (80) & 53 (59) & 58 (64) \\
+ PCA & 63 (70) & 47 (53) & 50 (56) & 62 (74) & 48 (54) & 53 (57) \\
+ SAE & 64 (72) & 49 (56) & 53 (59) & 64 (72) & 50 (57) & 55 (57) \\
DAS & \textbf{75} (\textbf{85}) & \textbf{57} (\textbf{67}) & \textbf{62} (\textbf{70}) & \textbf{75} (\textbf{83}) & \textbf{58} (\textbf{64}) & \textbf{63} (\textbf{70}) \\
\midrule
$\Psi$ (ours) & \rev{68 (80)} & \rev{54 (60)} & \rev{59 (64)} & \rev{64 (77)} & \rev{54 (63)} & \rev{60 (67)} \\
\bottomrule
\end{tabular}
\par\bigskip
\begin{tabular}[t]{@{}lcccc@{}}
\toprule
 & \multicolumn{4}{c}{IOI} \\
\cmidrule(l){2-5}
 & \multicolumn{2}{c}{MSE $\downarrow$} & \multicolumn{2}{c}{IIA (\%) $\uparrow$} \\
\cmidrule(lr){2-3}\cmidrule(l){4-5}
Method & $S_{\mathrm{Pos}}$ & $S_{\mathrm{Tok}}$ & $S_{\mathrm{Pos}}$ & $S_{\mathrm{Tok}}$ \\
\midrule
\nul{No Change} & \nul{16.19} & \nul{5.39} & \nul{33.3} & \nul{100.0} \\
\nul{Random} & \nul{5.87} & \nul{3.69} & \nul{40.5} & \nul{99.4} \\
\nul{Full Vector} & \nul{2.45} & \nul{2.82} & \nul{77.2} & \nul{95.7} \\
DBM & 2.22 & 2.35 & --- & --- \\
+ PCA & 2.24 & 2.33 & --- & --- \\
DAS & \textbf{2.20} & \textbf{2.08} & --- & --- \\
\midrule
DAS (reproduced) & 2.27 & 2.93 & 79.0 & 98.8 \\
$\Psi$ (ours) & 2.29 & 3.26 & \textbf{80.1} & 99.7 \\
\bottomrule
\end{tabular}
\end{table}

\paragraph{RAVEL.} The RAVEL task \citep{huang2024ravel} queries a language model about the continent, country and language of a city, and asks a localization method to change one attribute while leaving the others alone. A base prompt queries one attribute at a time, so an intervention on an attribute the prompt is not asking about is supposed to change nothing\rev{. We score it with the protocol of the published baselines \citep{mueller2025mib}
: a test set in which half of the pairs query the intervened attribute, two generated tokens and the benchmark's answer checker. $\Psi$ trails DAS in all six cells of }Table~\ref{tab:ravel}  \rev{and leads the other published methods in all but one. On this task the model's reading over the library, which lists the values of all three attributes, is often diffuse: for continent, its top entry is the gold value on $43.8\%$ of prompts}.

\paragraph{IOI.} This is the one task whose transport rule is supplied by the benchmark's high-level model rather than measured from the subject, as Appendix~\ref{sec:appendix-programs} sets out, and we report it on those terms. Its subject is \model{gpt2-small}. Indirect object identification \citep{wang2023interpretability} asks a model to complete a sentence such as \texttt{"When Mary and John went to the store, John gave a drink to"} with the name that appeared once rather than twice. MIB derives two variables from it: $S_{\mathrm{Pos}}$, the position of the repeated subject, and $S_{\mathrm{Tok}}$, its token identity. This task is scored differently from every other one in the benchmark. Rather than asking whether the intervened model produces the expected answer, it measures:

\begin{equation*}
  \mathbb{E}_{\iota}\big[\,\big(\Delta_{\phi}(\iota) - \Delta_{\mathbf{H}}(\iota)\big)^{2}\,\big],
  \qquad
  \Delta = \operatorname{logit}(\texttt{IO}) - \operatorname{logit}(\texttt{S}),
\end{equation*}

i.e., the mean-squared error between the logit difference the intervened subject produces and the one the high-level model $\mathbf{H}$ prescribes; lower is better. The abstraction site is also fixed: the alignment is carried by designated attention heads previously identified by \citet{wang2023interpretability} (three for $S_{\mathrm{Pos}}$ and one for $S_{\mathrm{Tok}}$) rather than by
a swept layer.

Table~\ref{tab:ravel} (bottom) reports our results. Under the objective used everywhere else in this paper, $\Psi$ lags behind the official DAS results and the other \rev{trained }baselines: it reaches $\PcqSpositionVFourMse$ on $S_{\mathrm{Pos}}$ and $\PcqStokenVFourMse$ on $S_{\mathrm{Tok}}$. We also ran our own reproduction of the DAS method, which reaches $\DasSpositionMse$ on $S_{\mathrm{Pos}}$, close to the official $\IoiPublishedDasPositionMse$, but $\DasStokenMse$ on $S_{\mathrm{Tok}}$, substantially higher than the official $\IoiPublishedDasTokenMse$.

Two readings of those numbers are worth separating. The first is that the two variables are not equally decisive, and both metrics say so. The MSE has no natural zero, so it is read between its two nulls. On $S_{\mathrm{Pos}}$ doing nothing costs $\IoiNochangePositionMse$, a random subspace of the matched rank $\IoiRandomPositionMse$, and every trained method between $\IoiTrainedPositionMseLow$ and $\IoiTrainedPositionMseHigh$. On $S_{\mathrm{Tok}}$ the same three anchors are $\IoiNochangeTokenMse$, $\IoiRandomTokenMse$ and $\IoiTrainedTokenMseLow$ to $\IoiTrainedTokenMseHigh$: the whole scale there is narrower than the stretch a random draw already covers on $S_{\mathrm{Pos}}$. Scored as an interchange accuracy on the same audits, doing nothing reaches $\PcqSpositionVFourFloor$\% and $\PcqStokenVFourFloor$\% respectively: moving the subject's position changes which name the model should emit, and moving its token identity does not. On the variable where the answer does move, $\Psi$ reaches $\PcqSpositionVFourNatural$\% against $\DasSpositionNatural$\% for our DAS reproduction, compared with $\PcqSpositionVFourRaw$\% for a whole-vector interchange at the same heads and $\IoiRandomPositionIia$\% for a random subspace. On the other, every method except the whole-vector interchange, doing nothing included, sits within a point of the ceiling, and only the logit difference can be measured at all, which is precisely the
reason the benchmark authors opted for the MSE metric on this task.

The second concerns what the objective aims at. A reading over two single-token names expresses the benchmark's margin exactly, since $\log (q_{\texttt{IO}} / q_{\texttt{S}}) = \operatorname{logit}(\texttt{IO}) - \operatorname{logit}(\texttt{S})$, and the transport table we build for IOI is precisely such a margin passed through a sigmoid. What differs is the loss and where its numbers come from: we minimize a cross-entropy against that target while the benchmark reports a squared error on the margin itself, so the two weight deviations differently, and the target is supplied by the benchmark's linear model rather than measured from the subject (Appendix~\ref{sec:appendix-programs}). We therefore read the distance between $\PcqStokenVFourMse$ and the published $\IoiPublishedDasTokenMse$ as the cost of optimizing a different quantity under a supervision the rest of this paper does not use. The distance between the same published entry and our own reproduction of the method that produced it, $\DasStokenMse$, should be read alongside it.

\subsection{Depthwise localization accuracy}
\label{sec:appendix-profiles}

\begin{figure}[ht]
\centering
\includegraphics[width=\textwidth]{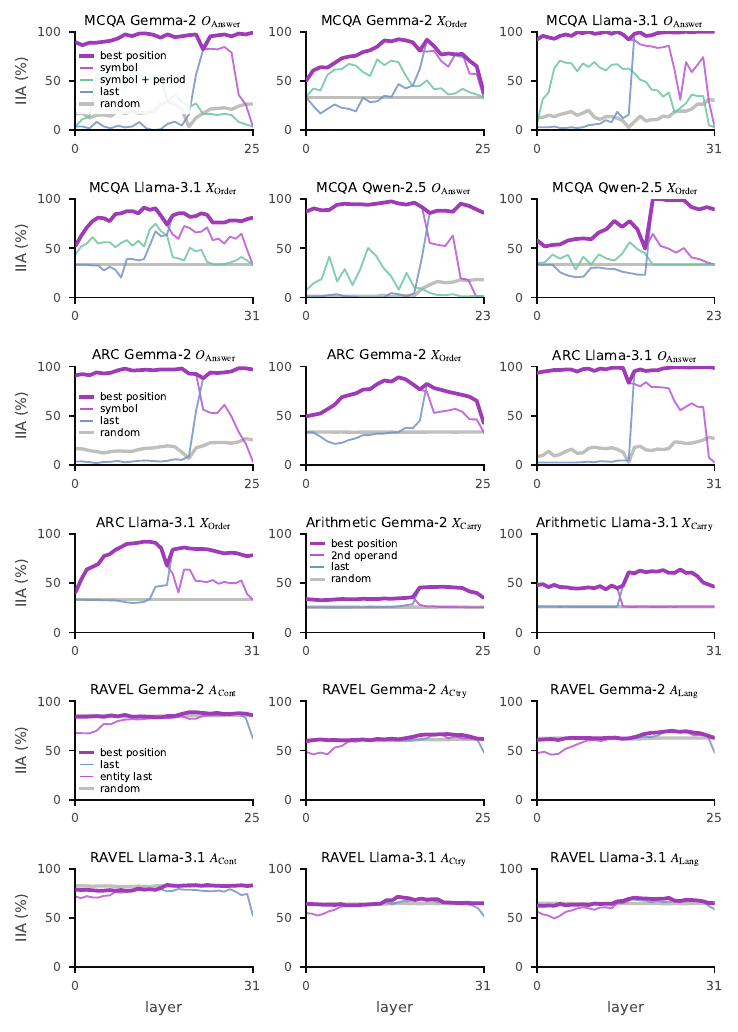}
\caption{\textbf{Depthwise localization accuracy.} Interchange accuracy of $\Psi$ at every layer of every cell, one thin line per token position and the thick line their maximum at each layer. The thick line is what Table~\ref{tab:localisation} summarizes: its average over layers is the depth mean and its highest point is the best layer. The gray line is the random subspace of the same rank, read the same way, wherever we have
measured it.}
\label{fig:depth}
\end{figure}

In this section we report for completeness the full depthwise intervention accuracy profiles for the different MIB tasks. The official benchmark scores reported in Table~\ref{tab:localisation} in fact only report two numbers per cell, i.e. mean and maximum. Figure~\ref{fig:depth} gives the profile they are computed from, which illustrates patterns that the aggregate score hides. On MCQA and ARC the token positions specialize with depth: an early layer is read best at the answer symbol and a late one at the last token, and the envelope is high across most of the network because one position or another carries the variable at each depth. On RAVEL the envelope tracks the random subspace along its whole length.

\subsection{Environment and compute}
\label{sec:appendix-compute}

All experiments are implemented in PyTorch (2.12) and evaluated by the benchmark's own evaluator at a pinned commit, in its own pinned environment, so our scores and the published baselines are produced by the same code path.

The localization results sweep every layer of each subject against the token positions the benchmark defines for that task, three for MCQA and two for ARC, arithmetic and RAVEL, which is \rev{$\LocalisationRunsLow$ to $\LocalisationRunsHigh$ }training runs per cell and \rev{$\LocalisationRuns$ }in total. Training is one run per site, for 32 epochs on MCQA, two on ARC and one on arithmetic and RAVEL, matching the budgets under which the baselines were tuned. These ran on H100 cards; the random floors, which need no training, and the control and supervision experiments ran on three RTX~A4000 cards. Cumulative compute for the project, including preliminary experiments that do not appear here, amounted to approximately 200 GPU-hours.

\subsection{Training hyperparameters}
\label{sec:appendix-hparams}

Every run in this paper trains the carrier with AdamW (weight decay $10^{-4}$, gradient norm clipped at $1$) and keeps $A$ orthonormal by taking the QR factor of an unconstrained matrix. The frozen subject runs in half precision, except \model{gpt2-small}, which runs in full precision\rev{; }\model{llama-3.1\_8b} \rev{is the instruction-tuned checkpoint (}\texttt{\rev{Meta-Llama-3.1-8B-Instruct}}\rev{) the benchmark specifies}. Single-seed results use seed $11$ and three-seed results use seeds $11$ to $13$. Each batch draws its sources by a fresh random derangement among prompts of the same transport group, and the carrier rank is the official DAS rank of each cell (Section~\ref{sec:results}). Table~\ref{tab:hparams} lists what differs between experiments; every value is read from the frozen configuration of the corresponding runs, which the supplementary material ships.

\begin{table}[ht]
\centering
\footnotesize
\setlength{\tabcolsep}{4pt}
\caption{Training settings. Localization runs use $\lambda = 0$ in Eq.~\eqref{eq:full_loss}. Control runs use $\lambda = 1$ and the gain controller of Eq.~\eqref{eq:linear_controller}, trained with learning rate $0.1$ against requests drawn half from other prompts' readings and half uniformly from the simplex, with \rev{an edit budget $c$ (Section~\ref{sec:control}) that every arm at a site shares. It is the 95th percentile of the distance between the activations of randomly paired training prompts, measured over the whole activation }at the MCQA site \rev{($c = \ControlCap$) and inside the frozen $\Psi$ carrier }at the Arithmetic site (\rev{$c = \ArithControlCap$, against $\ArithControlCapAmbient$ over the whole activation}). Where a row lists several batch sizes, they follow the order of its subjects.}
\label{tab:hparams}
\begin{tabularx}{\textwidth}{@{}l>{\raggedright\arraybackslash}Xcccl@{}}
\\
\toprule
Experiment & Subjects & Epochs & Batch & Learning rate & Mass mode \\
\midrule
\multicolumn{6}{@{}l}{\emph{Localization, every layer and benchmark position (Tables~\ref{tab:localisation} and~\ref{tab:ravel})}} \\
MCQA & \model{gemma-2\_2b}, \model{llama-3.1\_8b}, \model{qwen-2.5\_0.5b} & 32 & 16, 8, 32 & $3 \times 10^{-4}$ & \textsc{conditional} \\
ARC (Easy) & \model{gemma-2\_2b}, \model{llama-3.1\_8b} & 2 & 16 & $3 \times 10^{-4}$ & \textsc{conditional} \\
Arithmetic & \model{gemma-2\_2b}, \model{llama-3.1\_8b} & 1 & 8 & $10^{-3}$ & \textsc{full} \\
RAVEL & \model{gemma-2\_2b}, \model{llama-3.1\_8b} & 1 & 32, 16 & $10^{-3}$ & \textsc{conditional} \\
IOI, fixed heads & \model{gpt2-small} & 2 & 32 & $10^{-3}$ & \textsc{conditional} \\
\midrule
\multicolumn{6}{@{}l}{\emph{Supervision ablation (Figure~\ref{fig:supervision})}} \\
MCQA $X_{\mathrm{Order}}$, layer 17 & \model{gemma-2\_2b} & 32 & 16 & $3 \times 10^{-4}$ & \textsc{conditional} \\
\midrule
\multicolumn{6}{@{}l}{\emph{Control (Figure~\ref{fig:control})}} \\
MCQA $X_{\mathrm{Order}}$, layer 17 & \model{gemma-2\_2b} & 32 & 16 & $3 \times 10^{-4}$ & all three \\
Arithmetic $X_{\mathrm{Carry}}$, layer 21 & \model{gemma-2\_2b} & 1 & 8 & $10^{-3}$ & \textsc{conditional}, \textsc{full} \\
\bottomrule
\end{tabularx}
\end{table}

\end{document}